\documentclass{article} 
\usepackage{my_conference,times}

\usepackage{amsmath,amsfonts,bm}

\def\eqref#1{equation~\ref{#1}}

\def\1{\bm{1}}

\DeclareMathAlphabet{\mathsfit}{\encodingdefault}{\sfdefault}{m}{sl}
\SetMathAlphabet{\mathsfit}{bold}{\encodingdefault}{\sfdefault}{bx}{n}

\DeclareMathOperator*{\argmin}{arg\,min}

\usepackage{hyperref}
\usepackage{url}

\usepackage{booktabs}
\usepackage{multirow}
\usepackage{bm}
\usepackage{mathtools}
\usepackage{amsthm}
\newtheorem*{thm*}{Theorem}
\newtheorem*{prop*}{Proposition}
\usepackage{subcaption}

\newcommand{\defeq}{\vcentcolon=}
\usepackage{array}
\usepackage{wrapfig}

\title{Seeking Flat Minima with Mean Teacher on Semi- and Weakly-Supervised Domain Generalization for Object Detection}

\author{Ryosuke Furuta \& Yoichi Sato \\
Institute of Industrial Science\\
The University of Tokyo\\
Tokyo, Japan \\
\texttt{\{furuta,ysato\}@iis.u-tokyo.ac.jp}
}

\myfinalcopy 
\begin{document}

\maketitle

\begin{abstract}
Object detectors do not work well when domains largely differ between training and testing data.
To overcome this domain gap in object detection without requiring expensive annotations, we consider two problem settings: semi-supervised domain generalizable object detection (SS-DGOD) and weakly-supervised DGOD (WS-DGOD).
In contrast to the conventional domain generalization for object detection that requires labeled data from multiple domains, SS-DGOD and WS-DGOD require labeled data only from one domain and unlabeled or weakly-labeled data from multiple domains for training.
In this paper, we show that object detectors can be effectively trained on the two settings with the same Mean Teacher learning framework, where a student network is trained with pseudo-labels output from a teacher on the unlabeled or weakly-labeled data.
We provide novel interpretations of why the Mean Teacher learning framework works well on the two settings in terms of the relationships between the generalization gap and flat minima in parameter space.
On the basis of the interpretations, we also show that incorporating a simple regularization method into the Mean Teacher learning framework leads to flatter minima.
The experimental results demonstrate that the regularization leads to flatter minima and boosts the performance of the detectors trained with the Mean Teacher learning framework on the two settings.
\end{abstract}

\section{Introduction}
\label{sec:intro}


Object detection has been attracting much attention because it has practically useful applications such as in autonomous driving.
Object detectors have performed tremendously well on commonly used benchmark datasets for object detection, such as MSCOCO~\citep{lin2014microsoft} and PASCAL VOC~\citep{everingham2010pascal}.
However, such performance significantly drops when they are deployed on unseen domains, i.e., when the training and testing domains are different.
For example, Inoue et al.~\citep{inoue2018cross} reported a performance drop caused by the difference in image styles, and~Li et al.~\citep{li2022cross} showed one caused by the weather and time difference in the images captured with car-mounted cameras.

To solve this problem, many researchers have been exploring unsupervised domain adaptive object detection (UDA-OD)~\citep{deng2021unbiased,li2022cross,pmlr-v162-chen22b}.
On UDA-OD, we train object detectors using source domain data with ground-truth labels (bounding boxes and class labels) and unlabeled target domain data to adapt the detectors to the target domain.
However, in the real world, target domain data cannot always be accessed in the training phase.

Domain generalizable object detection (DGOD) is another common problem setting for solving the problem of the performance drop caused by the domain gaps~\citep{lin2021domain, zhang2022towards_dgod}.
On DGOD, we train object detectors using labeled data from multiple domains so that the detectors work well on unseen domains.
However, it is labor-intensive to collect these data for object detection because both bounding boxes and class labels for all objects in the images must be annotated.
Although single-DGOD~\citep{wu2022single,fan2023towards,vidit2023clip,wang2021robust,wang2023generalized,lee2024object}, on which we train object detectors to generalize unseen domains using labeled data from one single domain, has been investigated, the performance gain is still limited.

In this paper, we tackle two tasks as more realistic settings: i) semi-supervised DGOD (SS-DGOD)~\citep{malakouti2023semi} and ii) weakly-supervised DGOD (WS-DGOD).
The goal of SS-DGOD is to generalize object detectors to unseen domains using labeled data only from one single domain and unlabeled data from multiple domains. 
Note that the target domain data are not included in the training data.
On WS-DGOD, we use weakly labeled data from multiple domains instead of the unlabeled data in SS-DGOD.
``Weakly labeled'' means that we have only image-level labels that show the existence of each class in each training image and do not have bounding box annotations.
To the best of our knowledge, this is the first attempt to tackle WS-DGOD.
We show that object detectors can be effectively trained on the two settings with the same Mean Teacher learning framework, where a student network is trained with pseudo-labels output from a teacher on the unlabeled or weakly labeled data, and the teacher network is updated as the exponential moving average (EMA) of the student.

Not only do we experimentally demonstrate the good performance of the Mean Teacher learning framework, but also provide novel interpretations of why the Mean Teacher learning framework works well on these two settings in terms of the relationship between generalization ability and flat minima in parameter space.
These interpretations are based on our findings that the two key components of the Mean Teacher learning framework, i) EMA update and ii) learning from pseudo-labels, lead to flat minima during the training.
In the research area of domain generalization, it has been shown both theoretically and empirically that neural networks with flatter minima in parameter space have better generalization ability to unseen domains~\citep{foret2021sharpness, chaudhari2016entropy, cha2021swad, izmailov2018averaging, caldarola2022improving, wang2023sharpness, kaddour2022when, zhang2023flatness}.

On the basis of the interpretations, we also show that incorporating a simple regularization method into the Mean Teacher learning framework leads to flatter minima. 
Specifically, because the teacher and the student have similar loss values around the flat minima, we introduce an additional loss term so that the output from the student network becomes similar to that from the teacher network.
The experimental results demonstrate that the detectors trained with the Mean Teacher learning framework perform well for unseen test domains on the two settings.
We show that the simple yet effective regularization leads to flatter minima and boosts the performance of those detectors.

It is noteworthy that our aim is not to propose an entirely new method or surpass the state-of-the-art methods.
Instead, our contributions are summarized as follows:
\begin{itemize}
    \item We show that object detectors can be effectively trained on the SS-DGOD and WS-DGOD settings with the same Mean Teacher learning framework.
    \item We provide interpretations of why the detectors trained with the Mean teacher learning framework achieve robustness to unseen test domains in terms of the flatness of minima in parameter space.
    \item On the basis of the interpretations, we introduce a simple regularization method into the Mean Teacher learning framework to achieve flatter minima.
    \item We are the first to tackle the WS-DGOD setting.
\end{itemize}

\section{Problem Settings}\label{sec:problem_setting}
We formally describe the two problem settings of SS-DGOD and WS-DGOD.
Their goal is to obtain object detectors that perform well on unseen target domain data $\mathcal{D}_t=\{X_t\}$, where $X_t$ is a set of images from the target domain.

On SS-DGOD, we have labeled data from a source domain $\mathcal{D}_{s_1}=\{(X_{s_1},B_{s_1},C_{s_1})\}$ and unlabeled data from multiple source domains $\mathcal{D}_{s_i}=\{X_{s_i}\}_{i=2}^{N_D}$ in the training phase.
Here, $X_{s_1}=\{x^j_{s_1}\}_{j=1}^{N_{s_1}}$ is a set of $N_{s_1}$ images from domain $s_1$.
$B_{s_1}=\{b_{s_1}^j\}_{j=1}^{N_{s_1}}$ and $C_{s_1}=\{c_{s_1}^j\}_{j=1}^{N_{s_1}}$ are the corresponding bounding boxes and object-class labels, respectively.
${s_i}$ is the $i$-th source domain, and $N_{\mathcal{D}}$ is the number of the source domains.
We assume that the data distributions differ between the domains, i.e., $P(X_{s_1}) \neq P(X_{s_2}) \neq \cdots P(X_{s_{N_D}}) \neq P(X_{t})$.

On WS-DGOD, we use labeled data from a source domain $\mathcal{D}_{s_1}=\{(X_{s_1},B_{s_1},C_{s_1})\}$ and weakly labeled data from multiple domains $\mathcal{D}_{s_i}=\{(X_{s_i},C_{s_i})\}_{i=2}^{N_D}$ for training.

\begin{wraptable}{r}[0pt]{0.53\textwidth}
\caption{Formal comparisons of SS-DGOD, WS-DGOD, and related problem settings. DGOD stands for domain generalizable object detection, and SS-DGOD and WS-DGOD are semi-supervised DGOD and weakly-supervised DGOD, respectively. UDA-OD is unsupervised domain adaptive object detection.}
\centering
{\scriptsize
\begin{tabular}{ccc@{}} \toprule
     task & train data & test data \\ \midrule
     Single-DGOD & $\mathcal{D}_{s_1}=\{(X_{s_1},B_{s_1},C_{s_1})\}$ & $\mathcal{D}_t=\{X_t\}$ \\ \cmidrule(lr){1-3}
     \multirow{2}{*}{\bf{SS-DGOD}} & $\mathcal{D}_{s_1}=\{(X_{s_1},B_{s_1},C_{s_1})\}$, & \multirow{2}{*}{$\mathcal{D}_t=\{X_t\}$} \\
      & $\mathcal{D}_{s_i}=\{X_{s_i}\}_{i=2}^{N_D} $ & \\ \cmidrule(lr){1-3}
     \multirow{2}{*}{\bf{WS-DGOD}} & $\mathcal{D}_{s_1}=\{(X_{s_1},B_{s_1},C_{s_1})\}$, & \multirow{2}{*}{$\mathcal{D}_t=\{X_t\}$} \\
      & $\mathcal{D}_{s_i}=\{(X_{s_i},C_{s_i})\}_{i=2}^{N_D} $ & \\ \cmidrule(lr){1-3}
     DGOD & $\mathcal{D}_{s_i}=\{(X_{s_i},B_{s_i},C_{s_i})\}_{i=1}^{N_D}$ & $\mathcal{D}_t=\{X_t\}$ \\ \cmidrule(lr){1-3}
     \multirow{2}{*}{UDA-OD} & $\mathcal{D}_{s_1}=\{(X_{s_1},B_{s_1},C_{s_1})\}$, & \multirow{2}{*}{$\mathcal{D}_t=\{X_t\}$} \\
      & $\mathcal{D}_{t}=\{X_{t}\}$ & \\ \cmidrule(lr){1-3}
     \multirow{2}{*}{WSDA-OD} & $\mathcal{D}_{s_1}=\{(X_{s_1},B_{s_1},C_{s_1})\}$, & \multirow{2}{*}{$\mathcal{D}_t=\{X_t\}$} \\
      & $\mathcal{D}_{t}=\{(X_{t}, C_t)\}$ & \\ \bottomrule
\end{tabular}
}
\label{tbl:task_comparison}
\end{wraptable}

Table~\ref{tbl:task_comparison} compares SS-DGOD and WS-DGOD with related problem settings (Single-DGOD, DGOD, and UDA-OD).
As discussed in Sec.~\ref{sec:intro}, DGOD requires labeled data from multiple domains $\mathcal{D}_{s_i}=\{(X_{s_i},B_{s_i},C_{s_i})\}_{i=1}^{N_D}$, but those data are sometimes hard to prepare due to the high annotation cost.
In contrast, SS-DGOD (or WS-DGOD) requires labeled data from one domain $\mathcal{D}_{s_1}=\{(X_{s_1},B_{s_1},C_{s_1})\}$ and unlabeled data $\mathcal{D}_{s_i}=\{X_{s_i}\}_{i=2}^{N_D}$ (or weakly labeled data $\mathcal{D}_{s_i}=\{(X_{s_i},C_{s_i})\}_{i=2}^{N_D}$), which are easier to obtain.
Therefore, SS-DGOD and WS-DGOD are more practical settings than DGOD.
By using those data, we aim to better generalize object detectors to the unseen target domain data $\mathcal{D}_t=\{X_t\}$ than on Single-DGOD.

Another related setting is weakly-supervised domain adaptive object detection (WSDA-OD), a.k.a., cross-domain weakly-supervised object detection~\citep{inoue2018cross,hou2021informative,xu2022h2fa,tang2023detr}, which requires weakly-labeled target data $\mathcal{D}_t=\{(X_t, C_t)\}$ for training.
Unlike on UDA-OD and WSDA-OD, we can train the detectors even when the unlabeled or weakly-labeled target domain data ($\mathcal{D}_t=\{X_t\}$ or $\mathcal{D}_t=\{(X_t, C_t)\}$) are not accessible.

\section{Related Work}

\subsection{Domain Generalization for Image Classification}
Many methods have been proposed for domain generalization on image classification tasks as summarized in recent survey papers~\citep{zhou2022domain,wang2022generalizing}.
Among a variety of domain generalization methods, finding flat minima is one of the most common approaches~\citep{foret2021sharpness, chaudhari2016entropy, cha2021swad, izmailov2018averaging, caldarola2022improving, wang2023sharpness, kaddour2022when, zhang2023flatness}.
Those studies empirically and theoretically showed that finding flat minima in parameter space results in a better generalization ability.
\cite{izmailov2018averaging} and~\cite{cha2021swad} demonstrated that empirical risk minimization (ERM) with stochastic gradient descent (SGD) converges to the vicinity of a flat minimum, and averaging the parameter weights over a certain number of training steps/epochs results in reaching the flat minimum.
Inspired by these findings, we reveal that the Mean Teacher learning framework leads to flat minima, and thus can obtain good generalization ability.

\subsection{Domain Generalization for Object Detection}
Domain generalization for object detection has not been widely explored, compared with image classification. 
\cite{lin2021domain} proposed a method for disentangling domain-specific and domain-invariant features by adversarial learning on both image-level and instance-level features for DGOD. 
\cite{liu2020towards} investigated DGOD in underwater object detection and proposed DG-YOLO. 
For Single-DGOD, \cite{wang2021robust} proposed a self-training method that uses the temporal consistency of objects in videos.
\cite{wu2022single} proposed a method for disentangling domain-invariant features by contrastive learning and self-distillation.
\cite{fan2023towards} proposed perturbing the channel statistics of feature maps, which can be interpreted as data augmentation of image styles to a variety of domains.
\cite{wang2023generalized} proposed a disentangle method on frequency space for object detection from unmanned aerial vehicles.
\cite{vidit2023clip} proposed an augmentation method using a pre-trained vision-language model (CLIP) with textual prompts.

Unlike the above methods, as discussed in Sects.~\ref{sec:intro} and~\ref{sec:problem_setting}, we tackle SS-DGOD (Semi-Supervised Domain Generalization for Object Detection) and a new problem setting called WS-DGOD (Weakly-Supervised Domain Generalization for Object Detection).
The most closely related to our work is \cite{malakouti2023semi}'s work. 
They tackled SS-DGOD and proposed a language-guided alignment method.
However, the limitation of their method is that it requires a backbone network that was pre-trained on vision-and-language tasks.
Our experiments show that the object detectors trained with the Mean Teacher learning framework and our regularization outperform their method when the same backbone is used.



\subsection{Semi-supervised Domain Generalization}
There are a few methods that use both labeled and unlabeled data for domain generalization (SSDG) on image classification~\citep{zhang2022towards,zhou2023semi}.
\cite{zhang2022towards} proposed an unsupervised pre-training method called DARLING, which performs contrastive learning on unlabeled images to obtain domain-irrelevant feature representation.
\cite{zhou2023semi} extended a semi-supervised learning method called FixMatch~\citep{sohn2020fixmatch} to SSDG.

In contrast to those studies, we tackle SSDG for object detection.
We also tackle the ``weakly-labeled'' setting (i.e., WS-DGOD), which has not been explored even for image classification.

\subsection{Mean Teacher Learning Framework}
Mean Teacher learning framework was originally proposed for semi-supervised image classification~\citep{tarvainen2017mean}.
Several studies have investigated the use of the Mean Teacher learning framework for a variety of tasks such as domain generalization on image classification~\citep{yang2021adversarial}, (in-domain) weakly-supervised object detection~\citep{wang2022omni}, (in-domain) semi-supervised object detection~\citep{mi2022active}, 
UDA-OD~\citep{deng2021unbiased,li2022cross,he2022cross,deng2023harmonious,kennerley2024cat}, and UDA for semantic segmentaion~\citep{araslanov2021self,wang2021consistency,hoyer2022daformer,zhang2021prototypical}. 
\cite{lee2023exploring} provided a theoretical analysis of the Mean Teacher learning framework on masked image modeling pretext tasks for semi-supervised image classification.
We show that the Mean Teacher learning framework also works well on different settings (SS-DGOD and WS-DGOD), provide their interpretations, and introduce a simple regularization method to lead to flatter minima.

\begin{wrapfigure}{r}[0pt]{0.35\textwidth}
    \centering
    \includegraphics[width=0.35\columnwidth]{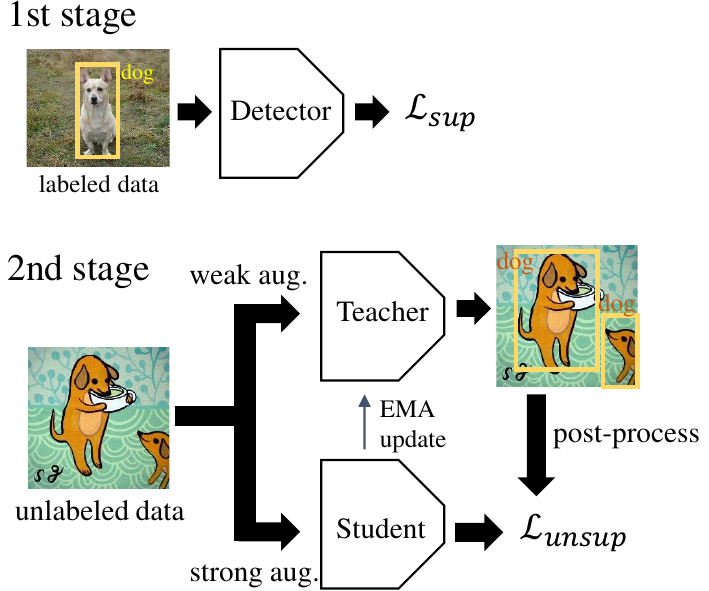}
    \caption{Training framework.}
    \label{fig:framework}
\end{wrapfigure}

\section{Training Method}\label{sec:training_method}
\subsection{Overview and Key Idea}\label{sec:method_overview}
On both SS-DGOD and WS-DGOD, our goal is to obtain object detectors that work well on the unseen target domain data $\mathcal{D}_t=\{X_t\}$.
\cite{gulrajanisearch} reported that if carefully implemented, empirical risk minimization (i.e., the image classifier simply trained with supervised learning on multiple domains) outperformed state-of-the-art domain generalization methods on several benchmark datasets for image classification.
Following this important finding, we expect similar behavior on object detection and aim to train an object detector on multiple domains $\mathcal{D}_{s_i} (i=1,\cdots, N_D)$.
However, we have no ground-truth labels (or have only weak labels) for $\mathcal{D}_{s_i} (i=2,\cdots, N_D)$ although ground-truth labels are available for $\mathcal{D}_{s_1}$.
Therefore, the question is how to train a detector on those domains.
Our solution is to use the Mean Teacher learning framework for object detection~\citep{li2022cross,pmlr-v162-chen22b} shown in Fig.~\ref{fig:framework}, where we have two networks (teacher and student) with the same structure and train the student network using the pseudo-labels generated by the teacher network.
Note that this Mean Teacher learning framework can be applied to any object detector, but we hereafter describe the loss functions of FasterRCNN~\citep{ren2015faster} as an example for ease of explanation.

\subsection{Pre-training}\label{sec:pretraining}
If we start the Mean Teacher learning from randomly initialized parameters, the teacher network cannot output reliable pseudo labels.
Therefore, we first perform supervised learning with the labeled data of one source domain $\mathcal{D}_{s_1}=\{(X_{s_1},B_{s_1},C_{s_1})\}$.
\begin{equation}
\begin{multlined}
    \mathcal{L}^{\mathrm{sup}}_{s_1}(\theta)=
    \mathcal{L}^{\mathrm{cls}}_{\mathrm{RPN}}(\theta, X_{s_1},B_{s_1},C_{s_1})+\mathcal{L}^{\mathrm{reg}}_{\mathrm{RPN}}(\theta, X_{s_1},B_{s_1},C_{s_1}) \\
    +\mathcal{L}^{\mathrm{cls}}_{\mathrm{RoI}}(\theta, X_{s_1},B_{s_1},C_{s_1})+\mathcal{L}^{\mathrm{reg}}_{\mathrm{RoI}}(\theta, X_{s_1},B_{s_1},C_{s_1}), \label{eq:loss_sup}
\end{multlined}
\end{equation}
where $\mathcal{L}^{\mathrm{cls}}_{\mathrm{RPN}}$ and $\mathcal{L}^{\mathrm{reg}}_{\mathrm{RPN}}$ are the classification and regression losses for region proposal networks (RPN), respectively.
$\mathcal{L}^{\mathrm{cls}}_{\mathrm{RoI}}$ and $\mathcal{L}^{\mathrm{reg}}_{\mathrm{RoI}}$ are those for RoIhead.
We initialize both the teacher and student networks with the parameters $\theta^* = \argmin_{\theta}\mathcal{L}^{\mathrm{sup}}_{s_1}(\theta)$ obtained from this pre-training.

\subsection{Mean Teacher Learning}\label{sec:student_teacher_learning}
\subsubsection{Generate Pseudo-labels}
Because we have no ground-truth labels (or have only weak labels) for the other source domains $\mathcal{D}_{s_i} (i=2,\cdots, N_D)$, we generate pseudo labels using the teacher network.
Specifically, we perform weak data augmentation to the unlabeled (or weakly-labeled) image $x_{s_i}^j$ and input it into the teacher network.
We denote the output from the teacher as $\{(\hat{b}_{s_i}^{jr},\hat{p}_{s_i}^{jr})\}_{r=1}^{N_R}$, where $\hat{b}_{s_i}^{jr}$ and $\hat{p}_{s_i}^{jr}$ are the predicted bounding box and class probabilities for the $r$-th region of interests (RoI) in the $j$-th image, respectively, and $N_R$ is the number of output RoIs.

In the case of SS-DGOD, we simply perform post-processing $f_{post}$ to $(\hat{b}_{s_i}^{jr},\hat{p}_{s_i}^{jr})$ and obtain the pseudo label $(\bar{b}_{s_i}^{jr},\bar{c}_{s_i}^{jr}) = f_{\mathrm{post}}(\hat{b}_{s_i}^{jr},\hat{p}_{s_i}^{jr})$.
Post-processing $f_{post}$ indicates a simple thresholding function if we use ``hard'' pseudo labels like~\citep{li2022cross} and indicates a sharpening function if we use ``soft'' pseudo labels like~\citep{pmlr-v162-chen22b}.

In the case of WS-DGOD, we perform the refinement process of applying the weak labels to the predicted class probabilities $\hat{p}_{s_i}^{jr}$ immediately before post-processing $f_{post}$ to obtain more accurate pseudo labels as follows:
\begin{equation}
    (\bar{b}_{s_i}^{jr},\bar{c}_{s_i}^{jr}) = f_{\mathrm{post}}(\hat{b}_{s_i}^{jr},\hat{p}_{s_i}^{jr}), \ \ \ \
    \hat{p}_{s_i}^{jr}(k)=
    \begin{cases}
    \hat{p}_{s_i}^{jr}(k) & \text{if} \ k \in c_{s_i}^j\\
    0 & \text{otherwise}
    \end{cases} \label{eq:refinement}
\end{equation}
where $\hat{p}_{s_i}^{jr}(k)$ is the predicted class probability for the $k$-th class.
Using the weak label $c_{s_i}^j$, Eq. (\ref{eq:refinement}) makes the predicted probability zero at each RoI if the $k$-th class does not exist in the $j$-th image.

\subsubsection{Update Student}
Now we have the pseudo labels $\bar{B}_{s_i}=\{\bar{b}_{s_i}^j\}_{j=1}^{N_{s_i}}$ and $\bar{C}_{s_i}=\{\bar{c}_{s_i}^j\}_{j=1}^{N_{s_i}}$ and train the student network with them.

We perform strong data augmentations to the image $x_{s_i}^j$ and input it into the student network.
In domain $s_1$, because the ground-truth labels are available, we update the student by backpropagating loss $\mathcal{L}^{\mathrm{sup}}_{s_1}$ in Eq. (\ref{eq:loss_sup}).
In the other domains $s_i (i=2,\cdots,N_D)$, we calculate loss $\mathcal{L}^{\mathrm{unsup}}_{s_i}$ using the pseudo labels and backpropagate it to update the student.
In summary, we update the parameters of student  $\theta^{\mathrm{student}}$ with loss $\mathcal{L}^{\mathrm{student}}$ as follows:
\begin{gather}
    \theta^{\mathrm{student}} \leftarrow \theta^{\mathrm{student}} - \nabla_{\theta} \mathcal{L}^{\mathrm{student}}(\theta), \ \ \ \
    \mathcal{L}^{\mathrm{student}}(\theta) = \mathcal{L}^{\mathrm{sup}}_{s_1}(\theta)+ \sum_{i=2}^{N_D}\mathcal{L}^{\mathrm{unsup}}_{s_i}(\theta)\label{eq:student_loss}\\
\begin{multlined}
    \mathcal{L}^{\mathrm{unsup}}_{s_i}(\theta)= 
    \shoveright{\mathcal{L}^{\mathrm{cls}}_{\mathrm{RPN}}(\theta, X_{s_i},\bar{B}_{s_i},\bar{C}_{s_i})+\mathcal{L}^{\mathrm{reg}}_{\mathrm{RPN}}(\theta, X_{s_i},\bar{B}_{s_i},\bar{C}_{s_i})} \\ 
    +\mathcal{L}^{\mathrm{cls}}_{\mathrm{RoI}}(\theta, X_{s_i},\bar{B}_{s_i},\bar{C}_{s_i})+\mathcal{L}^{\mathrm{reg}}_{\mathrm{RoI}}(\theta, X_{s_i},\bar{B}_{s_i},\bar{C}_{s_i}). \label{eq:loss_unsup}
\end{multlined}
\end{gather}

\subsubsection{Update Teacher}
Similar to previous studies~\citep{pmlr-v162-chen22b,li2022cross}, we do not update the parameters of the teacher $\theta^{\mathrm{teacher}}$ by backpropagation to obtain stable pseudo labels. Instead, we update them by the exponential moving average (EMA) of the parameters of the student network ${\theta^{\mathrm{teacher}} \leftarrow \alpha \theta^{\mathrm{teacher}}+(1-\alpha)\theta^{\mathrm{student}}}$.
Here, $\alpha$ is a hyperparameter to control the update speed.

\section{Why Does Mean Teacher Become Robust to Unseen Domains?}\label{sec:interpretation}

\begin{wrapfigure}{r}[0pt]{0.35\textwidth}
    \centering
    \includegraphics[width=0.35\columnwidth]{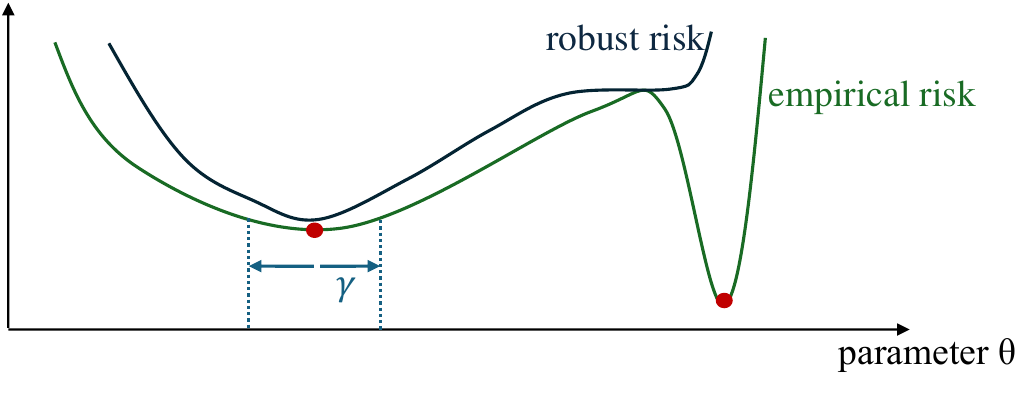}
    \caption{Empirical and robust risks.}
    \label{fig:RRM}
\end{wrapfigure}

We provide novel interpretations of why the Mean Teacher learning framework works well on SS-DGOD and WS-DGOD settings in terms of the relationship between generalization ability and flat minima in parameter space.
We show that the two key components of the Mean Teacher learning framework, i) EMA update and ii) learning from pseudo labels, lead to flat minima during the training.

\subsection{Definition}\label{sec:definition}
We define an empirical risk as $\mathcal{E}_\mathrm{ER}(\theta) \defeq \sum_{i=1}^{N_D}\mathcal{L}^\mathrm{sup}_{s_i}(\theta)$ when we assume that ground-truth labels are available on all the training domains. 
A risk at the target domain is defined as $\mathcal{E}_t(\theta)\defeq \mathcal{L}^{sup}_t(\theta)$.
The goal is to minimize the test risk $\mathcal{E}_t(\theta)$ by only solving the empirical risk minimization (ERM), i.e., $\min_{\theta}\mathcal{E}_\mathrm{ER}(\theta)$.
Hereafter, we use the terms {\it risk} and {\it loss} interchangeably.

\subsection{Preliminary Knowledge}
Previous studies for domain generalization demonstrated both theoretically and empirically that neural networks with flatter minima in parameter space exhibit superior generalization ability to unseen domains~\citep{foret2021sharpness, chaudhari2016entropy, cha2021swad, izmailov2018averaging, caldarola2022improving, wang2023sharpness, kaddour2022when, zhang2023flatness}.
\cite{cha2021swad} theoretically revealed the relationship between the flat minima and generalization gap (i.e., performance drop by domain shift).
We briefly describe the theorem for the subsequent explanation.
We consider the worst-case loss within neighbor regions in parameter space, which is defined as a robust risk $\mathcal{E}_{RR}^{\gamma}(\theta)\defeq \max_{{\|\Delta}\| \le \gamma}\mathcal{E}_{ER}(\theta+\Delta)$.
Here, $\gamma$ is the radius of the neighbor region.
As shown in Fig.~\ref{fig:RRM}, when $\gamma$ is sufficiently large, sharp minima of the empirical risk are not minima of the robust risk.
In contrast, the minima of the robust risk (i.e., $\argmin_{\theta} \mathcal{E}_{RR}^{\gamma}(\theta)$) are also minima in the flat regions of the empirical risk.
The following theorem shows the relationship between the optimal solution of robust risk minimization (RRM):
\begin{thm*}[from~\citep{cha2021swad}]
Consider a set of $N$ covers $\{\Theta_k\}_{k=1}^N$ such that the parameter space $\Theta \subset \cup_k^N\Theta_k$ where $\mathrm{diam}(\Theta)\defeq \mathrm{sup}_{\theta,\theta'\in\Theta}\|\theta-\theta'\|_2$, $N\defeq \lceil (\mathrm{diam}(\Theta)/\gamma)^d\rceil$ and $d$ is dimension of $\Theta$. Let $\theta^{\gamma}$ denote the optimal solution of the RRM, i.e., $\theta^{\gamma} \defeq \argmin_{\theta}\mathcal{E}_{RR}^{\gamma}(\theta)$, and let $v_k$ and $v$ be VC dimensions of each $\Theta_k$ and $\Theta$, respectively. Then, the gap between the optimal test loss, $\min_{\theta'}\mathcal{E}_{t}(\theta')$, and the test loss of $\theta^{\gamma}$, $\mathcal{E}_{t}(\theta^{\gamma})$, has the following bound with probability of at least $1-\delta$.
\begin{equation}
\begin{multlined}
    \mathcal{E}_{t}(\theta^{\gamma})-\min_{\theta'}\mathcal{E}_{t}(\theta') \le \mathcal{E}_{RR}^{\gamma}(\theta^{\gamma})-\min_{\theta''}\mathcal{E}_{ER}(\theta'')+\frac{1}{N_D}\sum_{i=1}^{N_D}\mathrm{Div}(s_i,t)\\
    +\max_{k\in[1,N]}\sqrt{\frac{v_k \ln(m/v_k)+\ln(2N/\delta)}{m}}+\sqrt{\frac{v\ln(m/v)+\ln(2/\delta)}{m}},\label{eq:theorem}
\end{multlined}
\end{equation}
where m is the number of training samples and $\mathrm{Div}(s_i,t)\defeq2\mathrm{sup}_A|\mathbb{P}_{s_i}(A)-\mathbb{P}_{t}(A)|$ is a divergence between two distributions. 
\end{thm*}

For its proof, see~\citep{cha2021swad}.
From the theorem, we can interpret that the gap between the RRM and ERM (i.e., $\mathcal{E}_{RR}^{\gamma}(\theta^{\gamma})-\min_{\theta''}\mathcal{E}_{ER}(\theta'')$) upper bounds the generalization gap in the test domain (i.e., $\mathcal{E}_{t}(\theta^{\gamma})-\min_{\theta'}\mathcal{E}_{t}(\theta')$).
Intuitively, as shown in Fig~\ref{fig:RRM}, the gap between the RRM and ERM narrows at flat regions of ERM.
Therefore, we can interpret that lowering the gap leads to flat minima of ERM and results in better generalization performance on the target domain.

\subsection{EMA Update}\label{sec:ema_update}
We explain why the EMA update in the Mean Teacher learning framework leads to flat minima.
\cite{stephan2017stochastic} showed that optimizing with constant SGD (i.e., SGD with a fixed learning rate) converges to a Gaussian distribution centered on the optimum.
On the basis of this finding, \cite{izmailov2018averaging} and \cite{cha2021swad} showed that the ERM with SGD converges to the marginal of a flat minimum, and averaging the weights of the parameters over some training steps/epochs leads to the flat minima.
To avoid overfitting, \cite{izmailov2018averaging} and \cite{cha2021swad} proposed sophisticated algorithms called SWA and SWAD for averaging the weights, and \cite{arpit2022ensemble} introduced a carefully designed averaging strategy called SMA.
In contrast to them, we found that a simple EMA also leads to flat minima, even without using those averaging algorithms.
This finding has not been provided in previous works, although the theoretical explanations for the benefit of averaging weights have already been provided.
The experiments presented in Sec.~\ref{sec:exp} show that the teacher network with only the EMA update of the student (i.e., without pseudo labeling) as shown in Eqs. (\ref{eq:ss-dgod_loss}-\ref{eq:ema}) can reach flatter minima and perform better than the student.
\begin{align}
    \theta^{\mathrm{student}} &\leftarrow \theta^{\mathrm{student}} - \nabla_{\theta} \mathcal{L}^{\mathrm{student}}(\theta), \ \ \ \ 
    \mathcal{L}^{\mathrm{student}}(\theta) = \mathcal{L}^{\mathrm{sup}}_{s_1}(\theta)\label{eq:ss-dgod_loss}\\
    \theta^{\mathrm{teacher}} &\leftarrow \alpha \theta^{\mathrm{teacher}}+(1-\alpha)\theta^{\mathrm{student}}, \label{eq:ema}
\end{align}

\subsection{Learning from Pseudo Labels}\label{sec:PL}

\begin{wrapfigure}{r}[0pt]{0.35\textwidth}
    \centering
    \includegraphics[width=0.35\columnwidth]{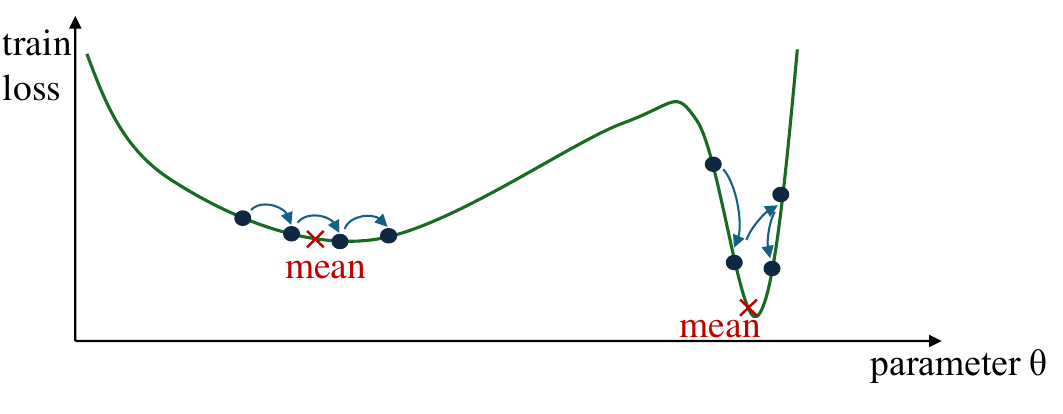}
    \caption{Intuitive interpretation of difference between loss values of trajectory of student and their mean (teacher).}
    \label{fig:mean_loss}
\end{wrapfigure}

We explain why learning from pseudo-labels in the Mean Teacher learning framework leads to flat minima.
Assuming that the pseudo-labels from the teacher are accurate enough (i.e., similar enough to ground truth), $\mathcal{L}_{s_i}^{unsup}$ in Eq. (\ref{eq:student_loss}) can be approximated by $\mathcal{L}_{s_i}^{sup}$, and we can regard the student network as the ERM in Sec.~\ref{sec:definition}.
On the other hand, as explained in Sec.~\ref{sec:ema_update} and shown in the experiments, because the teacher network updated with EMA has a better ability to reach flat minima than the student, the teacher can obtain less robust risk than the student, and we can regard the teacher as the robust risk minimizer.
Therefore, from Eq. (\ref{eq:theorem}), the smaller the difference between the losses of the teacher and student, the smaller the generalization gap in the target domain is.
Fig.~\ref{fig:mean_loss} shows its intuitive interpretations.
At the flat region, the trajectory of the student over the training steps and their mean (teacher) have similar loss values.
In contrast, there is a large difference between the loss values of the trajectory of the student and their mean at the sharp valley.

Next, we show that learning from pseudo-labels in the Mean Teacher learning framework makes the losses of the student and teacher similar.
Because the student is trained with the output from the teacher as pseudo-ground truth, the training promotes the outputs from the student similar to those from the teacher.
When we use monotonically increasing/decreasing functions with respect to the outputs as loss functions $\mathcal{E}$ (e.g., cross-entropy loss $\mathcal{E}(p)=p_{gt}\log(p)$), the more similar the outputs are, the more similar the loss values are, as shown below: 
\begin{prop*}
Assume $p_1<p_2<p_3\in \mathbb{R}$, and $\mathcal{E}(p): \mathbb{R}\rightarrow\mathbb{R}$ is a monotonically increasing/decreasing function of $p$.
Then, $|\mathcal{E}(p_3)-\mathcal{E}(p_2)|<|\mathcal{E}(p_3)-\mathcal{E}(p_1)|$ holds.
\end{prop*}
Let us consider $p_3$ as the teacher's output, and $p_2$ and $p_1$ as the outputs of the student. 
Since $p_2$ is closer to $p_3$ than $p_1$, the loss of $p_2$ becomes more similar to the loss of $p_3$ than that of $p_1$.
Therefore, we can interpret that learning from pseudo-labels align the outputs from the student to be similar to those from the teacher, thereby aligning the loss values, consequently leading to flat minima.

\section{Regularization for Flatter Minima}\label{sec:regul}

\subsection{Method}\label{sec:regularization_method}
As discussed in Sec.~\ref{sec:interpretation}, when the output from the student and teacher are similar, the networks tend to reach flat minima.
To this end, we introduce a simple regularization method to make the two networks' outputs more similar by training the student using raw outputs from the teacher.

\begin{wrapfigure}{r}[0pt]{0.35\textwidth}
    \centering
    \includegraphics[width=0.35\columnwidth]{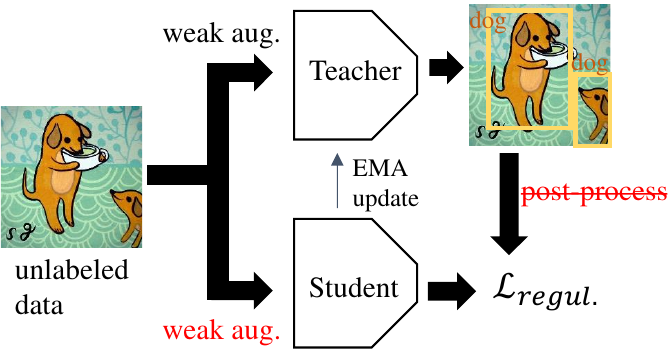}
    \caption{Overview of regualization method.}
    \label{fig:regul}
\end{wrapfigure}

Fig.~\ref{fig:regul} shows an overview of the method.
The concept is to apply regularization so that the outputs from the two networks are similar for the same input image.
Specifically, we perform weak data augmentations to the unlabeled (or weakly labeled) image $x_{s_i}^j$ and input the image into the teacher network.
We then use the output from the teacher $\{(\hat{b}_{s_i}^{jr},\hat{p}_{s_i}^{jr})\}_{r=1}^{N_R}$ directly as pseudo-ground truth {\it without post-processing} $f_{post}$.
To update the student, we input the {\it same weakly augmented image} $x_{s_i}^j$ into the student and calculate the regularization loss $\mathcal{L}^{\mathrm{regul.}}$ as follows:
\begin{gather}
    \mathcal{L}^{\mathrm{student}}(\theta) = \mathcal{L}^{\mathrm{sup}}_{s_1}(\theta)+ \sum_{i=2}^{N_D}[\mathcal{L}^{\mathrm{unsup}}_{s_i}(\theta)+\beta \mathcal{L}^{\mathrm{regul.}}_{s_i}(\theta)]\label{eq:student_loss_with_regul}\\
\begin{multlined}
    \mathcal{L}^{\mathrm{regul.}}_{s_i}(\theta)= 
    \shoveright{\mathcal{L}^{\mathrm{cls}}_{\mathrm{RPN}}(\theta, X_{s_i},\hat{B}_{s_i},\hat{C}_{s_i})+\mathcal{L}^{\mathrm{reg}}_{\mathrm{RPN}}(\theta, X_{s_i},\hat{B}_{s_i},\hat{C}_{s_i})} \\ 
    +\mathcal{L}^{\mathrm{cls}}_{\mathrm{RoI}}(\theta, X_{s_i},\hat{B}_{s_i},\hat{C}_{s_i})+\mathcal{L}^{\mathrm{reg}}_{\mathrm{RoI}}(\theta, X_{s_i},\hat{B}_{s_i},\hat{C}_{s_i}), \label{eq:loss_regul}
\end{multlined}
\end{gather}
where $\hat{B}_{s_i}=\{\hat{b}_{s_i}^j\}_{j=1}^{N_{s_i}}$ and $\hat{C}_{s_i}=\{\hat{c}_{s_i}^j\}_{j=1}^{N_{s_i}}$ are the {\it raw pseudo-labels} from the teacher, and $\beta$ is a hyperparameter to tune the strength of the regularization.

The differences between the regularization and the traditional Mean Teacher loss in Sec.~\ref{sec:student_teacher_learning} are 1) the use of weak augmentation instead of strong augmentation, and 2) the omission of post-processing (i.e., the sharpening function of \citep{pmlr-v162-chen22b} in our experiments). 
These approaches ensure that 1) the same input is given to both the student and teacher, and 2) the raw output from the teacher is used as pseudo-labels, which encourages closer alignment between the student and teacher.

\subsection{Connection to Prior Arts}
We can regard the regularization method as a type of knowledge distillation as the student is trained to mimic the raw output from the teacher.
Although the technical details are different, it has been empirically shown that knowledge distillation methods are effective on related tasks such as Single-DGOD~\citep{wu2022single}, domain adaptive semantic segmentation~\citep{zhang2021prototypical}, UDA-OD~\citep{cao2023contrastive, deng2023harmonious}, and semi-supervised domain adaptive object detection (where a small part of labeled target data $\mathcal{D}_t=\{(X_t, C_t)\}$ is accessible during the training~\citep{zhou2023ssda}).
We believe that our interpretation revealed one of the reasons knowledge-distillation methods lead to better generalization ability.

\section{Experiments}\label{sec:exp}
\subsection{Dataset Details}
We used the artistic style image dataset~\citep{inoue2018cross}, which has four domains: natural image, clipart, comic, and watercolor. 
The natural image domain has 16,551 images from PASCAL VOC07\&12, and the other domains have 1,000, 2,000, and 2,000 images, respectively. 
There are six object classes (bike, bird, car, cat, dog, and person), and we removed the images that do not contain these classes.

We conducted the experiments on three patterns of domains.
In the first pattern, we set the natural image domain as the labeled domain $s_1$ and set clipart and comic as the unlabeled domains $s_2, s_3$. 
We set watercolor as the target domain $t$. 
Concretely, we used the labeled trainval set of PASCAL VOC 2007\&2012, the unlabeled train set of clipart, and the unlabeled train set of comic for training. 
We then used the test sets of clipart and comic for validation. 
For evaluation (testing), we used the test set of watercolor. 
In the second and third patterns, we set $(s_1, s_2, s_3, t)=(\mathrm{natural, watercolor, comic, clipart})$ and $(s_1, s_2, s_3, t)=(\mathrm{natural, watercolor, clipart, comic})$, respectively.
The results on another dataset are shown in the supplementary material~\ref{sec:exp_car}.

\subsection{Implementation Details}\label{sec:implementation}
We used soft pseudo labeling proposed in~\citep{pmlr-v162-chen22b} for the Mean Teacher learning.
We used Gaussian FasterRCNN~\citep{pmlr-v162-chen22b} as the object detector, in which the regression output is modified to use the soft labels.
We used cross-entropy loss for both classification and regression losses, similar to~\citep{pmlr-v162-chen22b}.
We applied the same hyperparameters as in a previous study~\citep{pmlr-v162-chen22b} except for the number of iterations.
All training (including baseline models) was done with four A100 GPUs.
The parameters of the backbone network were initialized with the ResNet101 pre-trained on ImageNet.
The hyperparameters $\alpha$ in Eq. (\ref{eq:ema}) and $\beta$ in Eq. (\ref{eq:student_loss_with_regul}) were set to 0.9996 and 0.5 throughout the experiments, respectively.
During the inference (testing) phase, we used the teacher network.
Other details are given in the supplementary material.

\subsection{Baseline Methods}
As the baseline, we trained the detector {\it Gaussian FasterRCNN} on Single-DGOD setting (i.e., supervised learning on $s_1$ in Eq. (\ref{eq:loss_sup})).
To show the effectiveness of the EMA update, we trained {\it Gaussian FasterRCNN + EMA} with Eqs. (\ref{eq:ss-dgod_loss}-\ref{eq:ema}).
{\it Gaussian FasterRCNN + EMA + PL} is a detector trained with the Mean Teacher learning framework in Sec.~\ref{sec:training_method}.
{\it Gaussian FasterRCNN + EMA + PL + Regul.} is a detector with the Mean Teacher learning framework and the regualization in Eqs. (\ref{eq:student_loss_with_regul}-\ref{eq:loss_regul}).

To confirm the upper-bound performance, we also trained Gaussian FasterRCNN on DGOD and Oracle settings.
On DGOD, the detector was trained with supervised learning using the ground-truth labels on the domains $s_1, s_2$, and $s_3$.
On Oracle, the detector was trained with supervised learning on $s_1, s_2, s_3$, and the target domain $t$.

Because there is only one existing method on SS-DGOD (i.e., CDDMSL~\citep{malakouti2023semi}), we also compared the above detectors with state-of-the-art methods on related task settings such as Single-DGOD and UDA-OD.
It is noteworthy that existing DGOD methods such as~\citep{lin2021domain,liu2020towards} cannot be applied to SS-DGOD and WS-DGOD because they require labeled data from multiple source domains for training.
It is important to reiterate that our goal is not to propose a new method that outperforms state-of-the-art methods. 
Instead, our goal is to offer novel interpretations of the Mean Teacher and demonstrate that introducing simple regularization can lead to flatter minima, resulting in better robustness to unseen domains.

\subsection{Comparisons with Other Methods}

\begin{table*}[t]
\caption{Comparisons of mAP50 on the artistic style image dataset~\citep{inoue2018cross} when the target domain is watercolor.  Values with * are from previous study~\citep{li2022cross}.}
\centering
{\scriptsize
  \begin{tabular}{lllccc} \toprule
     \multirow{2}{*}{setting} & \multirow{2}{*}{method} & \multirow{2}{*}{backbone} & \multicolumn{3}{c}{mAP50} \\ \cmidrule(l){4-6}
      & & & watercolor & clipart & comic \\ \toprule
      Single-DGOD &  CLIP-based augmentation~\citep{vidit2023clip} & Res101 & 46.6 & 27.2 & {\bf 31.4} \\
      Single-DGOD & Gaussian FasterRCNN & Res101 & 50.5 & 34.5 & 26.6 \\
      Single-DGOD & Gaussian FasterRCNN + EMA & Res101 & {\bf 55.5} & {\bf 38.0} & 29.0 \\ \bottomrule
      SS-DGOD & CDDMSL~\citep{malakouti2023semi} & Res50 (RegionCLIP) & 46.1 & 39.1 & {\bf 38.3} \\
      SS-DGOD & CDDMSL~\citep{malakouti2023semi} & Res101 & 41.3 & 26.0 & 28.8 \\
      SS-DGOD & Gaussian FasterRCNN + EMA + PL & Res101 & 56.6 & 39.8 & 30.1 \\
      SS-DGOD & Gaussian FasterRCNN + EMA + PL + Regul. & Res101 & {\bf 58.2} & {\bf 43.3} & 32.2 \\ \bottomrule
      WS-DGOD & Gaussian FasterRCNN + EMA + PL & Res101 & 59.7 & 44.2 & 39.9 \\
      WS-DGOD & Gaussian FasterRCNN + EMA + PL + Regul. & Res101 & {\bf 62.9} & {\bf 46.2} & {\bf 40.2} \\ \bottomrule
      DGOD & Gaussian FasterRCNN & Res101 & 62.6 & 47.1 & 45.2 \\ \bottomrule
      Oracle & Gaussian FasterRCNN & Res101 & 62.2 & 48.2 & 48.6 \\ \bottomrule
      UDA-OD & Gaussian FasterRCNN + EMA + PL~\citep{pmlr-v162-chen22b} & Res101 & 54.9 & 43.4 & 27.0 \\
      UDA-OD & Gaussian FasterRCNN + EMA + PL + Regul. & Res101 & 58.8 & 45.4 & 32.7 \\
      UDA-OD & SCL*~\citep{shen2019scl} & Res101 & 55.2 & - & - \\
      UDA-OD & SWDA*~\citep{saito2019strong} & Res101 & 53.3 & - & - \\
      UDA-OD & UMT*~\citep{deng2021unbiased} & Res101 & 58.1 & - & - \\
      UDA-OD & AT*~\citep{li2022cross}  & Res101 & 59.9 & - & - \\
 \bottomrule
  \end{tabular} 
}
\label{tbl:comparison_watercolor_clipart_comic}
\end{table*}

Table~\ref{tbl:comparison_watercolor_clipart_comic} shows the results on the artistic image style dataset.
We evaluated with the mean average precision (mAP50) when the IoU threshold was 0.5.
EMA increased the mAP of {\it Gaussian FasterRCNN} from (50.5, 34.5, 26.6) to (55.5, 38.0, 29.0), and this was further boosted to (56.6, 39.8. 30.1) with pseudo labeling (PL).
We observed additional improvement to (58.2, 43.3, 32.2) with the regularization.
The regularization improved the performance not only on SS-DGOD but also on WS-DGOD.
The detectors trained on WS-DGOD performed better than those on SS-DGOD because WS-DGOD can generate more accurate pseudo labels by the refinement in Eq. (\ref{eq:refinement}).
Those results are comparable to those of the detectors trained on DGOD and Oracle.

For fair comparisons, we trained CDDMSL with Res101 backbone pre-trained on ImageNet.
However, its performance significantly degraded, as reported in a previous study~\citep{malakouti2023semi}, because it requires language-guided training, and initializing the model with RegionCLIP is crucial to achieve good performance.

The detectors trained on SS-DGOD and WS-DGOD also performed comparably to or better than those on UDA-OD, although we did not use the target domain data during the training.
Furthermore, the regularization can be directly applied to UDA-OD as well as SS-DGOD and WS-DGOD, and we also observed significant performance improvement by the regularization on UDA-OD.

\subsection{Analysis of Flatness}\label{sec:analysis_flatnes}
\begin{figure}[t]
\centering
  \begin{minipage}[b]{0.43\hsize}
    \centering
    \includegraphics[width=\columnwidth]{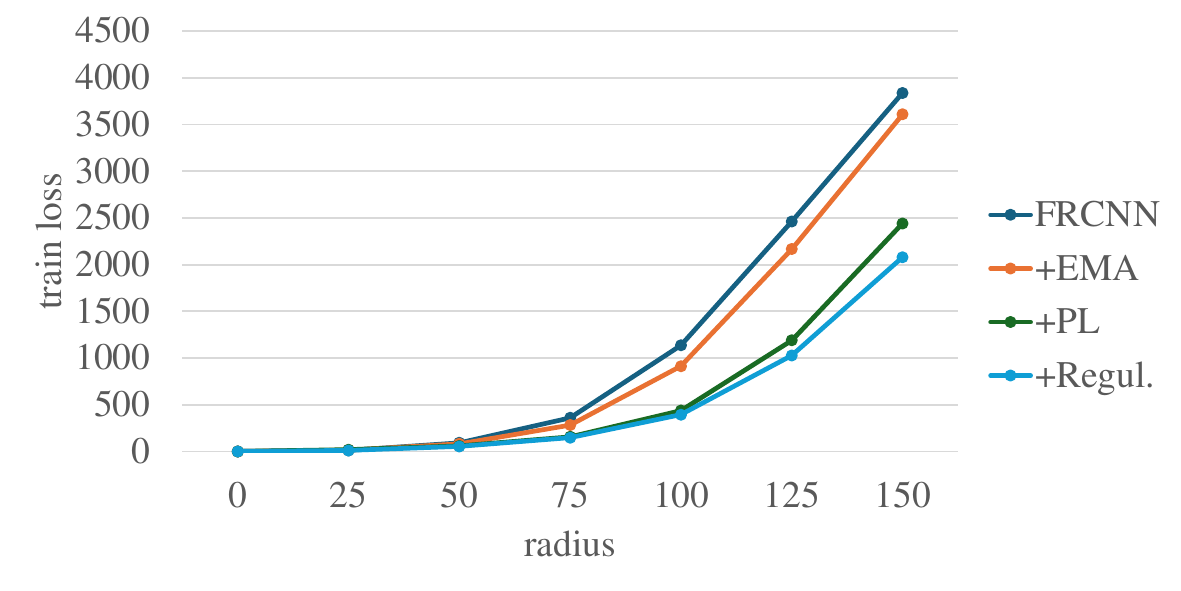}
  \end{minipage}
  \begin{minipage}[b]{0.43\hsize}
    \centering
    \includegraphics[width=\columnwidth]{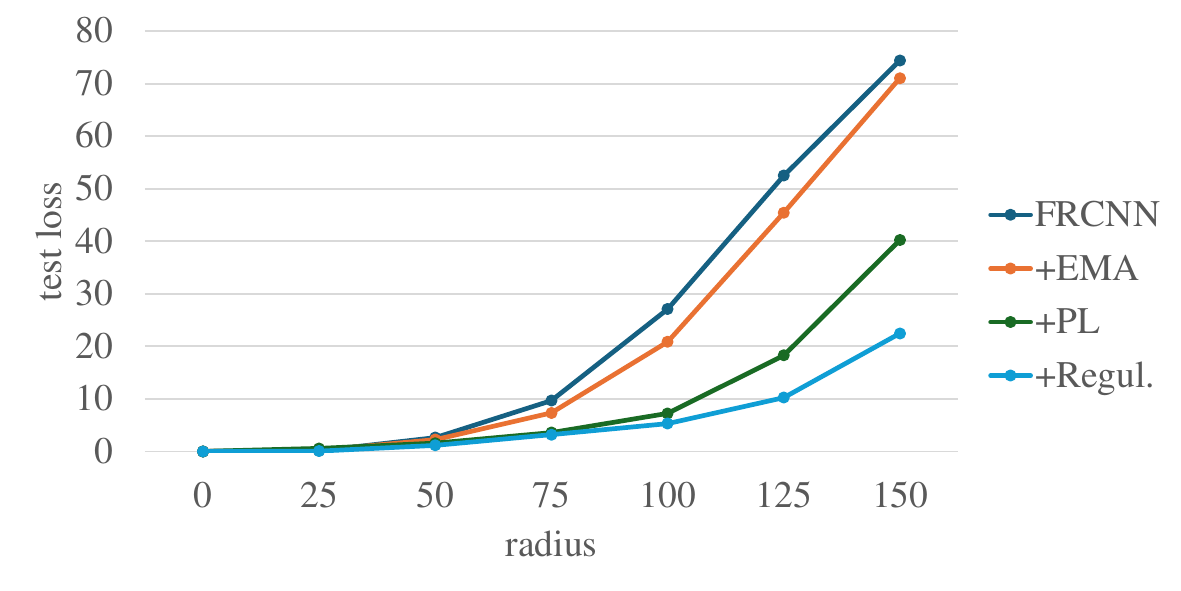}
  \end{minipage}
  \caption{Left and right plots compare average training and test flatness, respectively.}\label{fig:comp_flatness}
\end{figure}

To evaluate the flatness of the detectors in parameter space, following previous studies~\citep{izmailov2018averaging} and~\citep{cha2021swad}, we computed the change in loss values when we perturb the parameters.
Specifically, we sampled a random direction vector $d$ on a unit sphere, perturbed the parameters ($\theta'=\theta+d\gamma$) with a radius $\gamma$, and computed the average change over ten samples, i.e., $\mathcal{F}^{\gamma}(\theta)=\mathbb{E}_{\theta'}|\mathcal{E}(\theta')-\mathcal{E}(\theta)|$.
The lower the change is, the flatter the parameters.

Fig.~\ref{fig:comp_flatness} shows the $\mathcal{F}^{\gamma}(\theta)$ of the training loss $\mathcal{E}(\theta)=\sum_{i} \mathcal{L}_{s_i}^{sup}(\theta)$ and the test loss $\mathcal{E}(\theta)=\mathcal{L}_{t}^{sup}(\theta)$.
The training domains were $(s_1, s_2, s_3)$=(natural, watercolor, comic), and the test domain was clipart.
We can see that EMA, PL, and the regularization lowered the changes in the losses on both the training domains and test domain.
In other words, each contributed to falling into flatter minima.


\section{Conclusion and Limitation}\label{sec:conclusion}
We tackled two problem settings called semi-supervised domain generalizable object detection (SS-DGOD) and weakly-supervised DGOD (WS-DGOD) to train object detectors that can generalize to unseen domains.
We showed that the object detectors can be effectively trained on the two settings with the same Mean Teacher learning framework.
We also provided the interpretations of why the detectors trained with the Mean Teacher framework become robust to the unseen domains in terms of the flatness in the parameter space.
Based on the interpretations, we introduced a regularization method to lead to flatter minima, which makes the loss value of the student similar to that of the teacher.
The experiments showed that the detectors trained with the Mean Teacher learning framework and the regularization performed significantly better than the baseline methods.
Because Mean Teacher has been used across various tasks, our novel interpretation of why Mean Teacher becomes robust to unknown domains is likely to have a broad impact across a wide range of tasks.

The limitation is that the assumption in Sec.~\ref{sec:PL} does not always hold. 
Specifically, it is not always guaranteed that the pseudo labels from the teacher are accurate enough to approximate $\mathcal{L}_{s_i}^{unsup}$ with $\mathcal{L}_{s_i}^{sup}$. 
Nevertheless, we empirically showed that the Mean Teacher and the regularization lead to flatter minima in practice. 
There are two primary reasons for this observation.
First, when considering each domain independently, the assumption always holds in the labeled domain $s_1$, as labeled data is available, ensuring that $\mathcal{L}_{s_1}^{unsup} = \mathcal{L}_{s_1}^{sup}$. Second, the assumption is only necessary to explain how the Mean Teacher achieves flat minima in the {\it empirical risk} (i.e., the sum of the supervised losses $\mathcal{E}_{\mathrm{ER}}(\theta) = \sum_{i=1}^{N_D}\mathcal{L}^\mathrm{sup}_{s_i}(\theta)$).
Even if this assumption does not hold, we can similarly explain that the Mean Teacher reaches flat minima in the sum of supervised and unsupervised losses in Eq. (\ref{eq:student_loss}).
We believe that achieving flat minima in Eq. (\ref{eq:student_loss}) still positively affects robustness against unseen domains.
Further analysis of failure cases is left for future work.



\clearpage
\appendix
\section*{Appendix / Supplemental Material}

\section{More Analysis}
\subsection{How Sensitive to Hyperparameter $\beta$?}
Table~\ref{tbl:comparison_beta} shows the performance when the hyperparameter $\beta$ in Eq. (\ref{eq:student_loss_with_regul}) (i.e. strength of the regularization) was changed from 0 to 1.
By adding the regularization, the performance was constantly improved from the detector without regularization (i.e., $\beta=0$).

\begin{table}[h]
\caption{mAP50 with various $\beta$ on the artistic style image dataset~\citep{inoue2018cross}.}
\centering
{\scriptsize
  \begin{tabular}{llll} \toprule
     \multirow{2}{*}{setting} & \multirow{2}{*}{method} & \multirow{2}{*}{$\beta$} & mAP50 \\ \cmidrule(l){4-4}
      & & & clipart \\ \toprule
      SS-DGOD & Gaussian FasterRCNN + EMA + PL  & 0.0 & 39.8 \\
      SS-DGOD & Gaussian FasterRCNN + EMA + PL + Regul. & 0.25 & 40.7 \\
      SS-DGOD & Gaussian FasterRCNN + EMA + PL + Regul. & 0.5 & {\bf 43.3} \\
      SS-DGOD & Gaussian FasterRCNN + EMA + PL + Regul. & 0.75 & 42.1 \\
      SS-DGOD & Gaussian FasterRCNN + EMA + PL + Regul. & 1.0 & 42.5 \\
 \bottomrule
  \end{tabular} 
}
\label{tbl:comparison_beta}
\end{table}

\subsection{Importance of Encouraging Consistency}
\subsubsection{Comparison of Regularization with and without Post-processing}
In the regularization described in Sec.~\ref{sec:regularization_method}, we use the raw outputs from the teacher {\it without post-processing} to train the student so that the outputs from the two networks are similar.
To validate the claim, we compare the performance with and without post-processing (i.e., sharpening function~\citep{pmlr-v162-chen22b}) in the regularization in Eq. (\ref{eq:loss_regul}).
Table~\ref{tbl:comparison_post} shows that the performance drops when we perform the post-processing.
We observe that using raw outputs is important to obtain better performance.

\begin{table}[h]
\caption{mAP50 with and without post-processing on the artistic style image dataset~\citep{inoue2018cross}.}
\centering
{\scriptsize
  \begin{tabular}{llcl} \toprule
     \multirow{2}{*}{setting} & \multirow{2}{*}{method} & \multirow{2}{*}{post process} & mAP50 \\ \cmidrule(l){4-4}
      & & & clipart \\ \toprule
      SS-DGOD & Gaussian FasterRCNN + EMA + PL + Regul. & & {\bf 43.3} \\
      SS-DGOD & Gaussian FasterRCNN + EMA + PL + Regul. & \checkmark & 39.4 \\
 \bottomrule
  \end{tabular} 
}
\label{tbl:comparison_post}
\end{table}

\subsubsection{Importance of Consistent Augmentation between Teacher and Student}
\begin{figure*}[h]
    \centering
    \includegraphics[width=\linewidth]{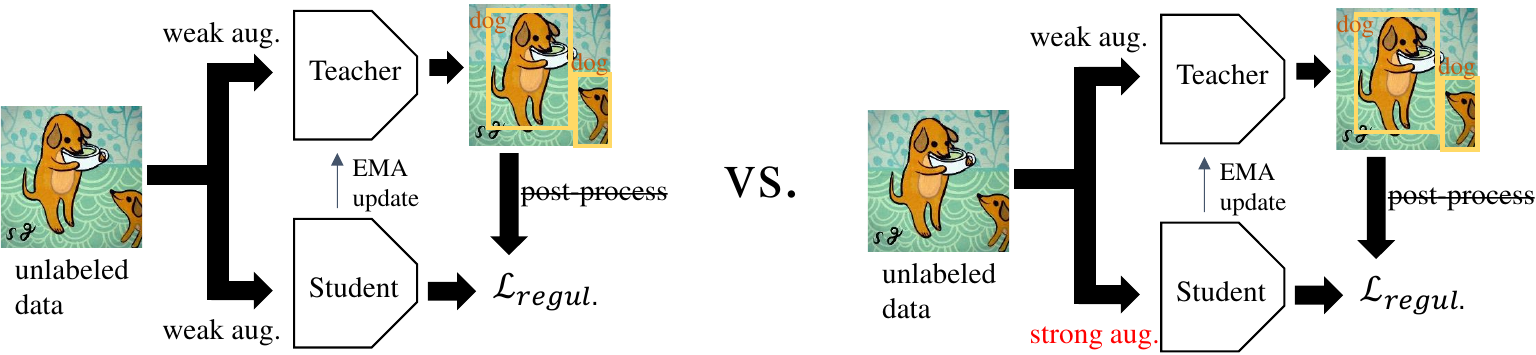}
    \caption{Comparisons between weak and strong augmentation for the student in the regularization.}
    \label{fig:comp_regul_consistency}
\end{figure*}

In the regularization described in Sec.~\ref{sec:regularization_method}, we input {\it weakly}-augmented images to the student (i.e., same input as the teacher) in order to encourage the consistency between the outputs from the teacher and student.
In this section, as shown in Fig.~\ref{fig:comp_regul_consistency}, we compare the weak and strong augmentation for the student in the regularization whereas weakly-augmented images were always input to the teacher.
Table~\ref{tbl:comparison_aug} shows that the weak augmentation obtains better performance than the strong augmentation, which implies the consistency between the outputs from the teacher and student using the same inputs leads to better performance.

\begin{table}[h]
\caption{Comparison of mAP50 between strong and weak augmentation in the regularization on the artistic style image dataset~\citep{inoue2018cross}.}
\centering
{\scriptsize
  \begin{tabular}{llcl} \toprule
     \multirow{2}{*}{setting} & \multirow{2}{*}{method} & \multirow{2}{*}{augmentation} & mAP50 \\ \cmidrule(l){4-4}
      & & & clipart \\ \toprule
      SS-DGOD & Gaussian FasterRCNN + EMA + PL + Regul. & weak & {\bf 43.3} \\
      SS-DGOD & Gaussian FasterRCNN + EMA + PL + Regul. & strong & 42.5 \\
 \bottomrule
  \end{tabular} 
}
\label{tbl:comparison_aug}
\end{table}

\subsection{Why is Only Weak Augmentation Used in the Regularization?}
\begin{figure*}[h]
    \centering
    \includegraphics[width=\linewidth]{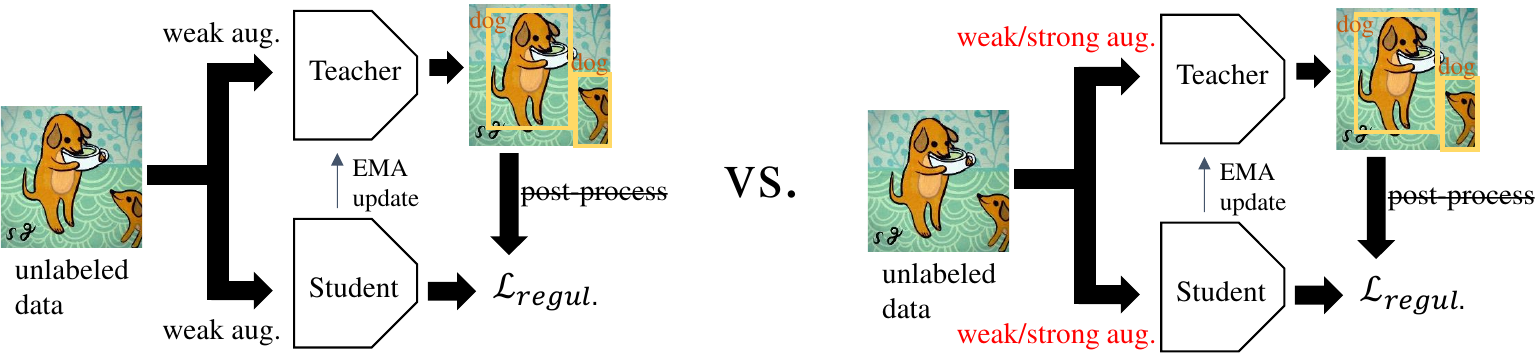}
    \caption{Comparisons between weak and random (weak/strong) augmentation in the regularization.}
    \label{fig:comp_random_aug}
\end{figure*}

One may think why only weak augmentation is used in the regualization in Fig.~\ref{fig:regul}, and what is the performance of randomly using strong and weak augmentation?
To answer this question, we evaluated the performance when randomly using strong and weak augmentation as shown in the right side of Fig.~\ref{fig:comp_random_aug}. 
In this setting, the strong and weak augmentation was randomly chosen with a probability of 0.5 at each iteration. 
When the strong augmentation was chosen, the same strongly augmented image was input into both the student and teacher networks. 
Then, the raw output from the teacher without post-processing was used to calculate the regularization loss for the students in Eqs. (\ref{eq:student_loss_with_regul}) and (\ref{eq:loss_regul}) to encourage consistency between the outputs from the student and teacher. 
As shown in Table~\ref{tbl:comparison_random_aug}, only weak augmentation obtained better performance. 
We think it is because inputting strongly augmented images into the teacher can make noisy pseudo-labels.

\begin{table}[h]
\caption{Comparison of mAP50 between weak and weak/strong (random) augmentation in the regularization on the artistic style image dataset.}
\centering
{\scriptsize
  \begin{tabular}{llcl} \toprule
     \multirow{2}{*}{setting} & \multirow{2}{*}{method} & \multirow{2}{*}{augmentation} & mAP50 \\ \cmidrule(l){4-4}
      & & & clipart \\ \toprule
      SS-DGOD & Gaussian FasterRCNN + EMA + PL & N/A & 39.8 \\
      SS-DGOD & Gaussian FasterRCNN + EMA + PL + Regul. & weak & {\bf 43.3} \\
      SS-DGOD & Gaussian FasterRCNN + EMA + PL + Regul. & weak/strong (random) & 41.1 \\
 \bottomrule
  \end{tabular} 
}
\label{tbl:comparison_random_aug}
\end{table}

\subsection{Class-wise Average Precision}
Table~\ref{tbl:comparison_watercolor_perclass},~\ref{tbl:comparison_clipart_perclass}, and~\ref{tbl:comparison_comic_perclass} show average precision (AP50) at each class when the target domain is watercolor, clipart, and comic, respectively.
We can see that the regularization improved the performance on many classes.

\begin{table}[h]
\caption{Comparisons of AP50 at each class on watercolor of the artistic style image dataset~\citep{inoue2018cross}. The values of * are from~\citep{li2022cross}.}
\centering
{\tiny
  \begin{tabular}{llwc{6mm}wc{6mm}wc{6mm}wc{6mm}wc{6mm}wc{6mm}wc{6mm}} \toprule
     setting & method & bicycle & bird & cat & car & dog & person & mAP \\ \toprule
      Single-DGOD & CLIP-based augmentation~\citep{vidit2023clip} & 74.8 & 37.3 & {\bf 36.8} & 40.7 & 29.2 & 59.9 & 46.4 \\
      Single-DGOD & Gaussian FasterRCNN & {\bf 90.4} & 47.9 & 30.3 & 46.7 & 28.7 & 59.2 & 50.5 \\
      Single-DGOD & Gaussian FasterRCNN + EMA & 86.2 & {\bf 54.3} & 35.3 & {\bf 53.5} & {\bf 34.5} & {\bf 69.0} & {\bf 55.5} \\ \bottomrule
      SS-DGOD & CDDMSL~\citep{malakouti2023semi} (RegionCLIP) & 66.3 & 50.6 & 34.5 & 49.2 & 20.1 & 56.0 & 46.1 \\
      SS-DGOD & CDDMSL~\citep{malakouti2023semi} (Res101) & 75.5 & 36.1 & 23.9 & 40.7 & 19.7 & 52.0 & 41.3 \\
      SS-DGOD & Gaussian FasterRCNN + EMA + PL & {\bf 87.4} & {\bf 54.6} & 40.0 & 51.9 & 32.4 & 73.1 & 56.6 \\
      SS-DGOD & Gaussian FasterRCNN + EMA + PL + Regul. & 87.2 & 52.3 & {\bf 44.7} & {\bf 53.2} & {\bf 36.8} & {\bf 75.3} & {\bf 58.2} \\\bottomrule
      WS-DGOD & Gaussian FasterRCNN + EMA + PL & 90.3 & 55.8 & 49.3 & 49.9 & 37.5 & 75.4 & 59.7 \\ 
      WS-DGOD & Gaussian FasterRCNN + EMA + PL + Regul. & {\bf 95.8} & {\bf 59.9} & {\bf 51.5} & {\bf 53.3} & {\bf 40.2} & {\bf 76.7} & {\bf 62.9} \\ \bottomrule
      DGOD & Gaussian FasterRCNN & 84.8 & 57.8 & 51.0 & 50.8 & 51.8 & 79.3 & 62.6 \\\bottomrule
      Oracle & Gaussian FasterRCNN & 90.9 & 59.9 & 44.2 & 53.1 & 46.7 & 78.3 & 62.2 \\ \bottomrule
      UDA-OD & Gaussian FasterRCNN + EMA + PL~\citep{pmlr-v162-chen22b} & 77.7 & 46.5 & 40.4 & 50.1 & 39.7 & 75.0 & 54.9 \\
      UDA-OD & Gaussian FasterRCNN + EMA + PL + Regul. & 82.8 & 51.4 & 43.2 & 59.3 & 39.0 & 77.0 & 58.8 \\
      UDA-OD & SCL*~\citep{shen2019scl} & 82.2 & 55.1 & 51.8 & 39.6 & 38.4 & 64.0 & 55.2 \\
      UDA-OD & SWDA*~\citep{saito2019strong} & 82.3 & 55.9 & 46.5 & 32.7 & 35.5 & 66.7 & 53.3 \\
      UDA-OD & UMT*~\citep{deng2021unbiased} & 88.2 & 55.3 & 51.7 & 39.8 & 43.6 & 69.9 & 58.1 \\
      UDA-OD & AT*~\citep{li2022cross}  & 93.6 & 56.1 & 58.9 & 37.3 & 39.6 & 73.8 & 59.9 \\
 \bottomrule
  \end{tabular} 
}
\label{tbl:comparison_watercolor_perclass}
\end{table}

\begin{table*}[h]
\caption{Comparisons of AP50 at each class on clipart of the artistic style image dataset~\citep{inoue2018cross}.}
\centering
{\tiny
  \begin{tabular}{llwc{6mm}wc{6mm}wc{6mm}wc{6mm}wc{6mm}wc{6mm}wc{6mm}} \toprule
     setting & method & bicycle & bird & cat & car & dog & person & mAP \\ \toprule
      Single-DGOD & CLIP-based augmentation~\citep{vidit2023clip} & 36.5 & 22.5 & {\bf 20.1} & 25.0 & 8.8 & 50.4 & 27.2 \\
      Single-DGOD & Gaussian FasterRCNN & 69.5 & 25.1 & 5.7 & {\bf 39.4} & 17.3 & 49.9 & 34.5 \\
      Single-DGOD & Gaussian FasterRCNN + EMA & {\bf 87.6} & {\bf 29.3} & 5.5 & 30.1 & {\bf 18.3} & {\bf 57.2} & {\bf 38.0} \\ \bottomrule
      SS-DGOD & CDDMSL~\citep{malakouti2023semi} (RegionCLIP) & 51.0 & {\bf 33.3} & {\bf 26.5} & {\bf 45.2} & 14.6 & 63.8 & 39.1 \\
      SS-DGOD & CDDMSL~\citep{malakouti2023semi} (Res101) & 41.6 & 19.2 & 5.5 & 26.7 & 12.3 & 50.9 & 26.0 \\
      SS-DGOD & Gaussian FasterRCNN + EMA + PL & 75.8 & 31.2 & 9.4 & 33.1 & 20.4 & {\bf 69.1} & 39.8 \\
      SS-DGOD & Gaussian FasterRCNN + EMA + PL + Regul. & {\bf 79.3} & 32.5 & 11.6 & 40.9 & {\bf 26.3} & 69.0 & {\bf 43.3} \\ \bottomrule
      WS-DGOD & Gaussian FasterRCNN + EMA + PL & 80.3 & {\bf 33.3} & 11.1 & {\bf 44.5} & {\bf 23.2} & {\bf 72.6} & 44.2 \\ 
      WS-DGOD & Gaussian FasterRCNN + EMA + PL + Regul. & {\bf 84.8} & 33.2 & {\bf 23.8} & 43.0 & 22.1 & 70.1 & {\bf 46.2} \\ \bottomrule
      DGOD & Gaussian FasterRCNN & 76.0 & 34.8 & 18.8 & 38.3 & 36.9 & 77.6 & 47.1 \\\bottomrule 
      Oracle & Gaussian FasterRCNN~ & 70.4 & 38.8 & 26.1 & 52.9 & 27.5 & 73.4 & 48.2 \\\bottomrule
      UDA-OD & Gaussian FasterRCNN + EMA + PL~\citep{pmlr-v162-chen22b} & 79.9 & 33.5 & 6.5 & 53.1 & 23.7 & 65.2 & 43.6 \\
      UDA-OD & Gaussian FasterRCNN + EMA + PL + Regul. & 72.4 & 35.4 & 16.0 & 57.2 & 19.7 & 71.5 & 45.4 \\
 \bottomrule
  \end{tabular} 
}
\label{tbl:comparison_clipart_perclass}
\end{table*}

\begin{table*}[h]
\caption{Comparisons of AP50 at each class on comic of the artistic style image dataset~\citep{inoue2018cross}.}
\centering
{\tiny
  \begin{tabular}{llwc{6mm}wc{6mm}wc{6mm}wc{6mm}wc{6mm}wc{6mm}wc{6mm}} \toprule
     setting & method & bicycle & bird & cat & car & dog & person & mAP \\ \toprule
      Single-DGOD & CLIP-based augmentation~\citep{vidit2023clip} & 29.0 & {\bf 18.6} & {\bf 27.6} & 32.7 & {\bf 28.4} & {\bf 52.2} & {\bf 31.4} \\
      Single-DGOD & Gaussian FasterRCNN & 45.0 & 10.8 & 9.5 & {\bf 33.8} & 17.5 & 43.0 & 26.6 \\
      Single-DGOD & Gaussian FasterRCNN + EMA & {\bf 50.0} & 15.0 & 11.2 & 26.8 & 22.4 & 48.3 & 29.0 \\ \bottomrule
      SS-DGOD & CDDMSL~\citep{malakouti2023semi} (RegionCLIP) & 41.8 & {\bf 27.8} & {\bf 23.5} & {\bf 44.2} & {\bf 34.8} & 57.8 & {\bf 38.3} \\
      SS-DGOD & CDDMSL~\citep{malakouti2023semi} (Res101) & {\bf 44.3} & 12.2 & 13.7 & 30.5 & 19.7 & 52.1 & 28.8 \\
      SS-DGOD & Gaussian FasterRCNN + EMA + PL & 41.5 & 14.5 & 11.4 & 24.5 & 27.3 & {\bf 61.1} & 30.1 \\
      SS-DGOD & Gaussian FasterRCNN + EMA + PL + Regul. & 42.3 & 15.6 & 15.9 & 31.5 & 30.2 & 57.8 & 32.2 \\ \bottomrule
      WS-DGOD & Gaussian FasterRCNN + EMA + PL & 53.7 & 23.1 & 19.9 & {\bf 44.3} & {\bf 33.7} & {\bf 64.5} & 39.9 \\ 
      WS-DGOD & Gaussian FasterRCNN + EMA + PL + Regul. & {\bf 54.2} & {\bf 23.2} & {\bf 23.8} & 44.1 & 31.5 & 64.2 & {\bf 40.2} \\ \bottomrule
      DGOD & Gaussian FasterRCNN & 54.6 & 29.5 & 33.5 & 38.9 & 43.2 & 71.4 & 45.2 \\\bottomrule 
      Oracle & Gaussian FasterRCNN~ & 55.7 & 29.0 & 44.5 & 46.3 & 45.1 & 71.3 & 48.6 \\\bottomrule
      UDA-OD & Gaussian FasterRCNN + EMA + PL~\citep{pmlr-v162-chen22b} & 42.2 & 13.6 & 10.8 & 16.6 & 19.3 & 59.5 & 27.0 \\
      UDA-OD & Gaussian FasterRCNN + EMA + PL + Regul. & 46.3 & 14.4 & 20.3 & 28.8 & 23.5 & 62.6 & 32.7 \\
 \bottomrule
  \end{tabular} 
}
\label{tbl:comparison_comic_perclass}
\end{table*}

\subsection{Qualitative Results}
\begin{figure*}[h]
    \begin{minipage}{\linewidth}
    \centering
    \includegraphics[width=\linewidth]{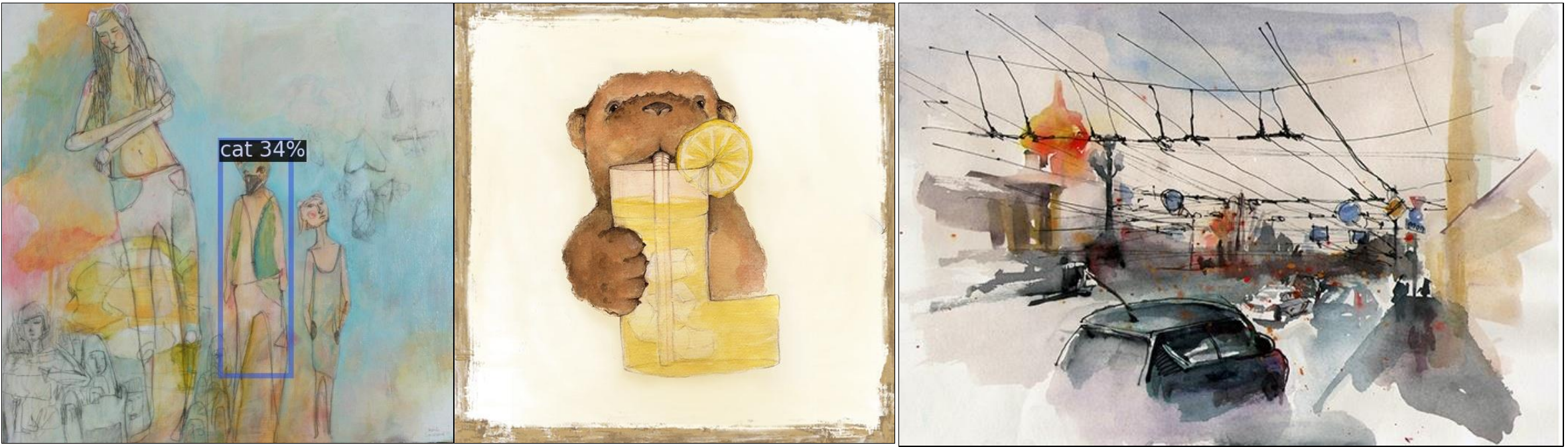}
    \subcaption{Gaussian FasterRCNN trained on Single-DGOD setting (i.e., trained with labeled data on PASCAL VOC07\&12).}
    \end{minipage}
    \begin{minipage}{\linewidth}
    \centering
    \includegraphics[width=\linewidth]{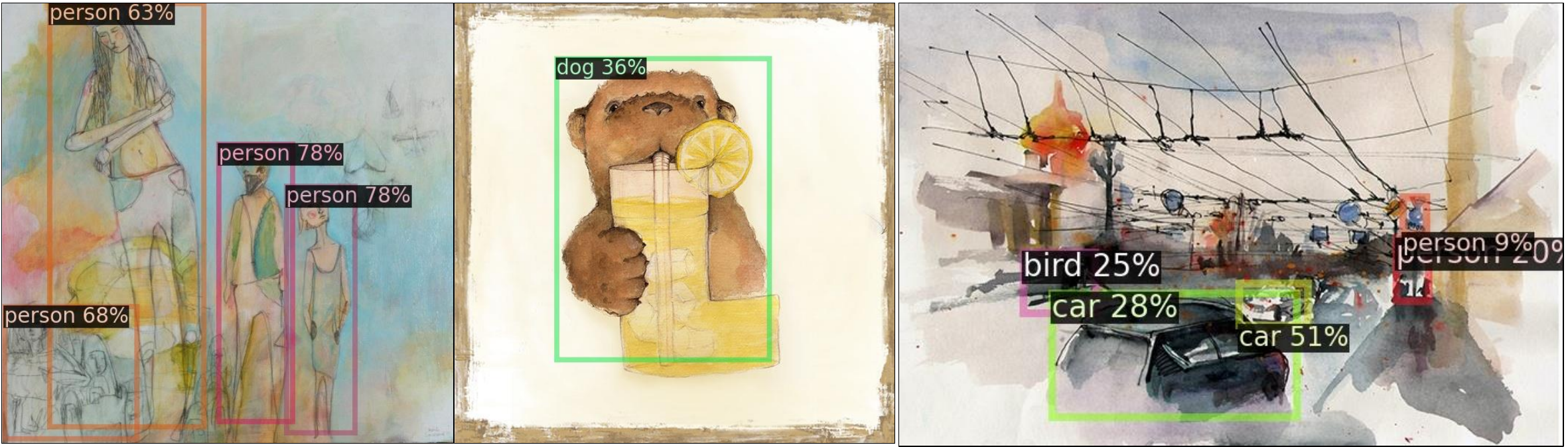}
    \subcaption{Gaussian FasterRCNN + EMA + PL + Regul. trained on SS-DGOD setting (i.e., trained with labeled data on PASCAL VOC07\&12 and unlabeled data on clipart and comic).}
    \end{minipage}
    \caption{Qualitative comparisons on watertcolor.}
    \label{fig:qualitative_comp_watercolor}
\end{figure*}

\begin{figure*}[h]
    \begin{minipage}{\linewidth}
    \centering
    \includegraphics[width=\linewidth]{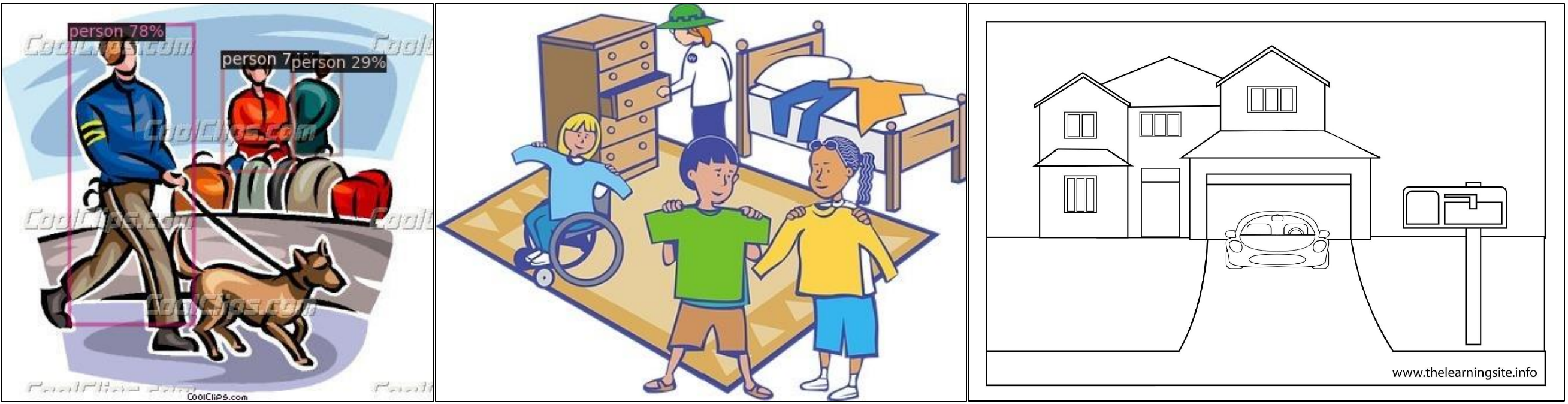}
    \subcaption{Gaussian FasterRCNN trained on Single-DGOD setting (i.e., trained with labeled data on PASCAL VOC07\&12).}
    \end{minipage}
    \begin{minipage}{\linewidth}
    \centering
    \includegraphics[width=\linewidth]{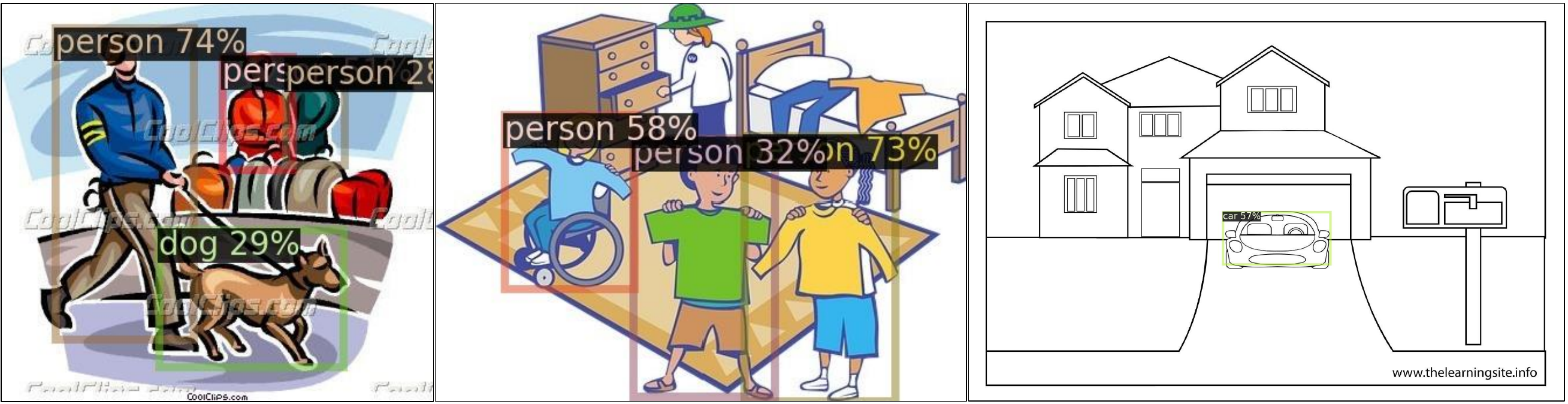}
    \subcaption{Gaussian FasterRCNN + EMA + PL + Regul. trained on SS-DGOD setting (i.e., trained with labeled data on PASCAL VOC07\&12 and unlabeled data on watercolor and comic).}
    \end{minipage}
    \caption{Qualitative comparisons on clipart.}
    \label{fig:qualitative_comp_clipart}
\end{figure*}

\begin{figure*}[h]
    \begin{minipage}{\linewidth}
    \centering
    \includegraphics[width=\linewidth]{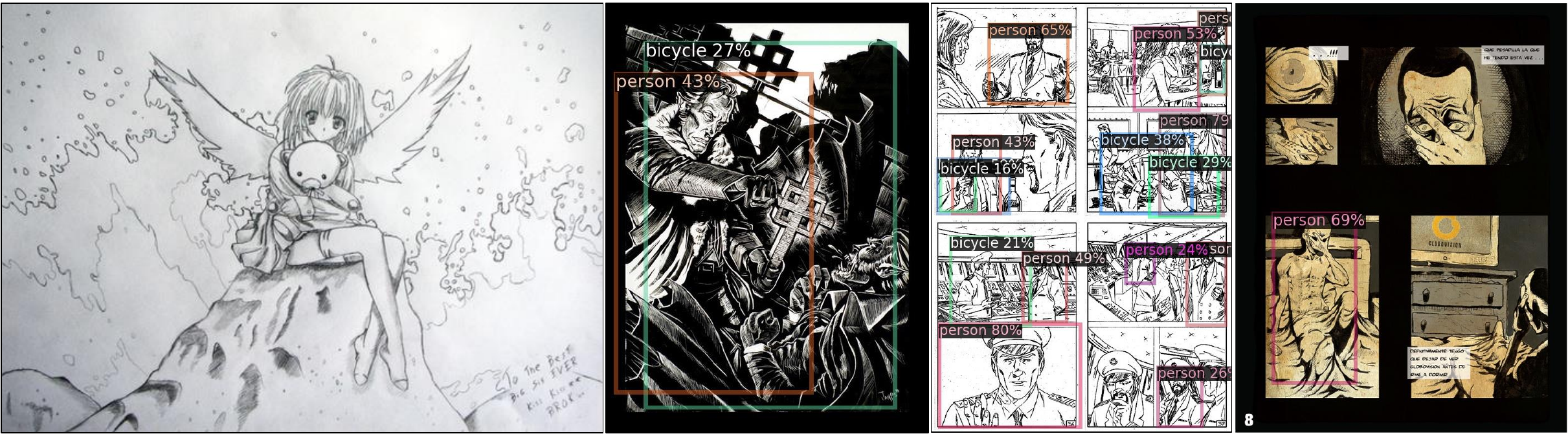}
    \subcaption{Gaussian FasterRCNN trained on Single-DGOD setting (i.e., trained with labeled data on PASCAL VOC07\&12).}
    \end{minipage}
    \begin{minipage}{\linewidth}
    \centering
    \includegraphics[width=\linewidth]{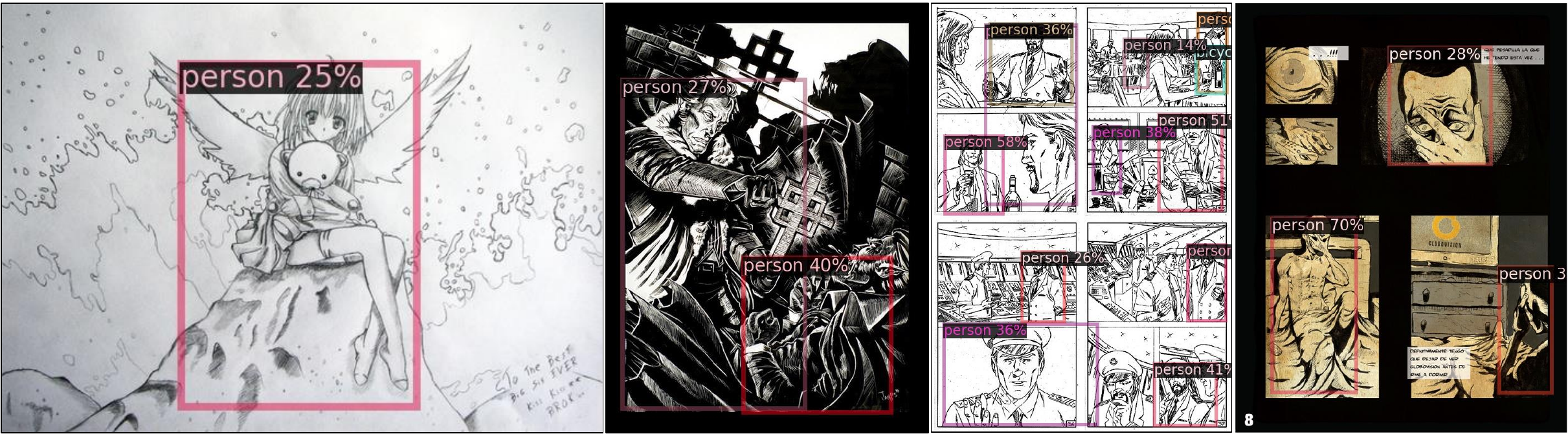}
    \subcaption{Gaussian FasterRCNN + EMA + PL + Regul. trained on SS-DGOD setting (i.e., trained with labeled data on PASCAL VOC07\&12 and unlabeled data on clipart and comic).}
    \end{minipage}
    \caption{Qualitative comparisons on comic.}
    \label{fig:qualitative_comp_comic}
\end{figure*}

Figs.~\ref{fig:qualitative_comp_watercolor}, \ref{fig:qualitative_comp_clipart}, and \ref{fig:qualitative_comp_comic} show the qualitative comparison on watercolor, clipart, and comic, respectively. 
We observe that false negative detection of the baseline model was drastically improved.

\subsection{When the Number of Unlabeled Domain is One}
In Sec.~\ref{sec:exp}, we conducted the experiments under the setting of one labeled domain and multiple unlabeled domains (e.g., $(s_1, s_2, s_3, t)$= (natural, clipart, comic, watercolor)).
In this section, we conducted the experiments with one labeled domain and one unlabeled domain, which are the same settings as the CDDSL paper~\citep{malakouti2023semi}: (s1, s2, t) = (natural, comic, watercolor) and (natural, comic, clipart).
Table~\ref{tbl:comparison_watercolor_clipart_one_unlabeled_domain} shows the superior performance to CDDMSL even on these settings.

\begin{table*}[h]
\caption{Comparisons of mAP50 on the artistic style image dataset when $(s_1, s_2, t)$= (natural, comic, watercolor) and (natural, comic, clipart).  Values with * are from previous study~\citep{malakouti2023semi}.}
\centering
{\scriptsize
  \begin{tabular}{lllcc} \toprule
     \multirow{2}{*}{setting} & \multirow{2}{*}{method} & \multirow{2}{*}{backbone} & \multicolumn{2}{c}{mAP50} \\ \cmidrule(l){4-5}
      & & & watercolor & clipart \\ \toprule
      SS-DGOD & CDDMSL*~\citep{malakouti2023semi} & Res50 (RegionCLIP) & 49.4 & 39.8 \\
      SS-DGOD & Gaussian FasterRCNN + EMA + PL & Res101 & 55.2 & 38.4 \\
      SS-DGOD & Gaussian FasterRCNN + EMA + PL + Regul. & Res101 & {\bf 56.5} & {\bf 40.1} \\ 
 \bottomrule
  \end{tabular} 
}
\label{tbl:comparison_watercolor_clipart_one_unlabeled_domain}
\end{table*}

\clearpage
\section{Results on Car-mounted Camera Dataset~\citep{wu2022single}}\label{sec:exp_car}
\subsection{Dataset Details}
The car-mounted camera dataset is a recently developed dataset in~\citep{wu2022single} for Single-DGOD or DGOD, where the images were selected from the standard datasets such as Cityscapes~\citep{cordts2016cityscapes}, FoggyCityscapes~\citep{sakaridis2018semantic}, BDD-100k~\citep{yu2020bdd100k}, and AdverseWeather~\citep{hassaballah2020vehicle}.
The domains were clearly redefined based on the weather and time differences: daytime-sunny, night-sunny, daytime-foggy, dusk-rainy, and night-rainy. 
The number of images for each domain is 27,708, 18,310, 2,642, 3,501, and 2,494, respectively. We used daytime-sunny as the labeled domain $s_1$ and used night-sunny and daytime-foggy as the unlabeled (or weakly-labeled) domains $s_2, s_3$. 
We used each of the remaining domains (dusk-rainy and night-rainy) as the target domain. 
Because the train/val/test split is not publicly available for daytime-sunny, dusk-rainy, and night-rainy, we used all images of daytime-sunny, the trainval set of night-sunny, and the trainval set of daytime-foggy for training. 
We then used the test set of night-sunny and the test set of daytime-foggy for validation. 
We used all images of dusk-rainy and night-rainy for evaluation (testing). 
There are seven object classes: bus, bike, car, motor, person, rider, and truck.

There are two reasons for using this dataset for evaluation. One is that the images in this dataset were selected from the standard datasets, and the other is that the domains were clearly redefined based on the weather and time differences as described above.
In the setting of the previous work~\citep{malakouti2023semi}, i.e., $(s_1, s_2, t)$ = (Cityscapes, FoggyCityscapes, BDD100k), the differences between domains are ambiguous. This is because Cityscape primarily assumes clear/medium daytime weather, Foggycityscape assumes foggy weather, while BDD100K includes various times of day and weather conditions.
Therefore, instead, we used the car-mounted camera dataset.

\subsection{Comparisons with Other Methods}

\begin{table*}[h]
\caption{Comparisons of mAP50 on the car-mounted camera dataset~\citep{wu2022single}. The values of * and ** were from~\citep{wu2022single} and~\citep{vidit2023clip}, respectively.}
\centering
{\scriptsize
  \begin{tabular}{llccc} \toprule
     \multirow{2}{*}{setting} & \multirow{2}{*}{method} & \multirow{2}{*}{backbone} & \multicolumn{2}{c}{mAP50} \\ \cmidrule(l){4-5}
      & & & dusk-rainy & night-rainy \\ \toprule
      Single-DGOD & FasterRCNN* & Res101  & 26.6 & 14.5 \\
      Single-DGOD & CDSD*~\citep{wu2022single} & Res101 & 28.2 & 16.6 \\
      Single-DGOD & CLIP-based augmentation**\citep{vidit2023clip} & Res101 & 32.3 & 18.7 \\
      Single-DGOD & Gaussian FasterRCNN & Res101 & 25.3 & 13.3 \\
      Single-DGOD & Gaussian FasterRCNN + EMA & Res101 & {\bf 36.0} & {\bf 19.0} \\ \bottomrule
      SS-DGOD & Gaussian FasterRCNN + EMA + PL & Res101 & 30.3 & 21.3 \\
      SS-DGOD & Gaussian FasterRCNN + EMA + PL + Regul. & Res101 & {\bf 31.2} & {\bf 21.9} \\ \bottomrule
      WS-DGOD & Gaussian FasterRCNN + EMA + PL & Res101 & 30.5 & 22.5 \\
      WS-DGOD & Gaussian FasterRCNN + EMA + PL + Regul. & Res101 & {\bf 32.5} & {\bf 23.1} \\ \bottomrule
      DGOD & Gaussian FasterRCNN & Res101 & 28.4 & 21.2 \\
 \bottomrule
  \end{tabular} 
}
\label{tbl:comparison_sdgod}
\end{table*}

Table~\ref{tbl:comparison_sdgod} shows the results on the car-mounted camera dataset.
Each of EMA, PL, and the regularization improved the performance on both target domains except that PL degraded the performance on dusk-rainy.
We will investigate the cause of the performance drop in our future work.

The mAP50 of the detector with the regularization is boosted to (32.5, 23.1) on WS-DGOD.
This result exceeds (32.3, 18.7), which is the result of CLIP-based augmentation~\citep{vidit2023clip} proposed for Single-DGOD.
Also, this result is better than those of the models trained with supervised learning on the three domains (DGOD).

\subsection{Analysis of Flatness}
\begin{figure}[h]
\centering
  \begin{minipage}[b]{0.32\hsize}
    \centering
    \includegraphics[width=\columnwidth]{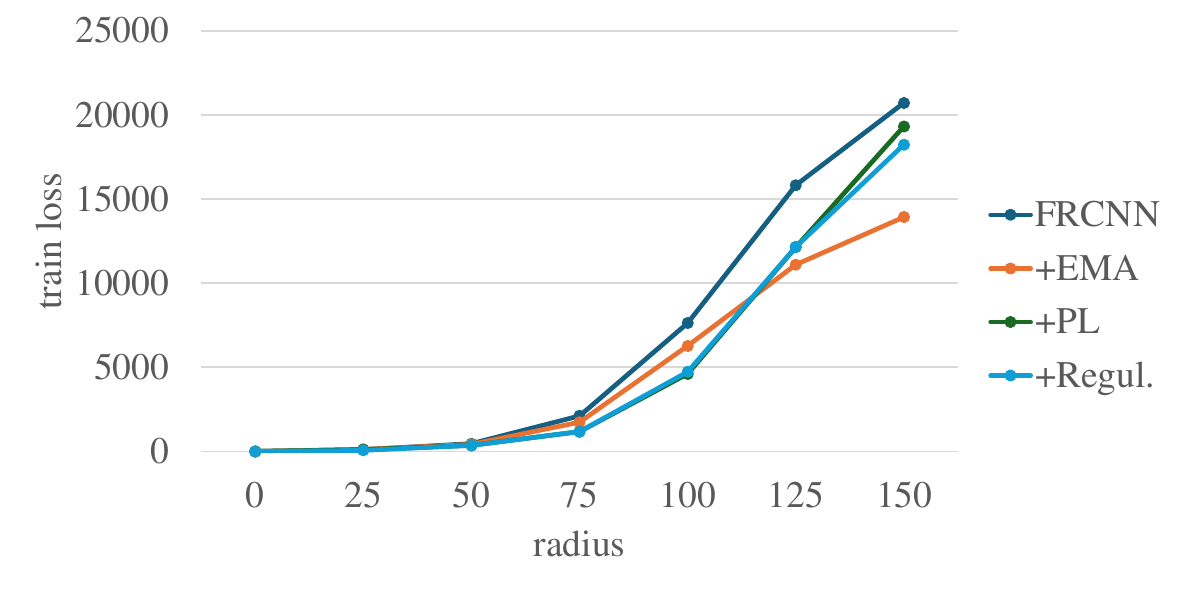}
    \subcaption{Daytime-clear}\label{fig:flatness_train_loss_daytime_clear}
  \end{minipage}
  \begin{minipage}[b]{0.32\hsize}
    \centering
    \includegraphics[width=\columnwidth]{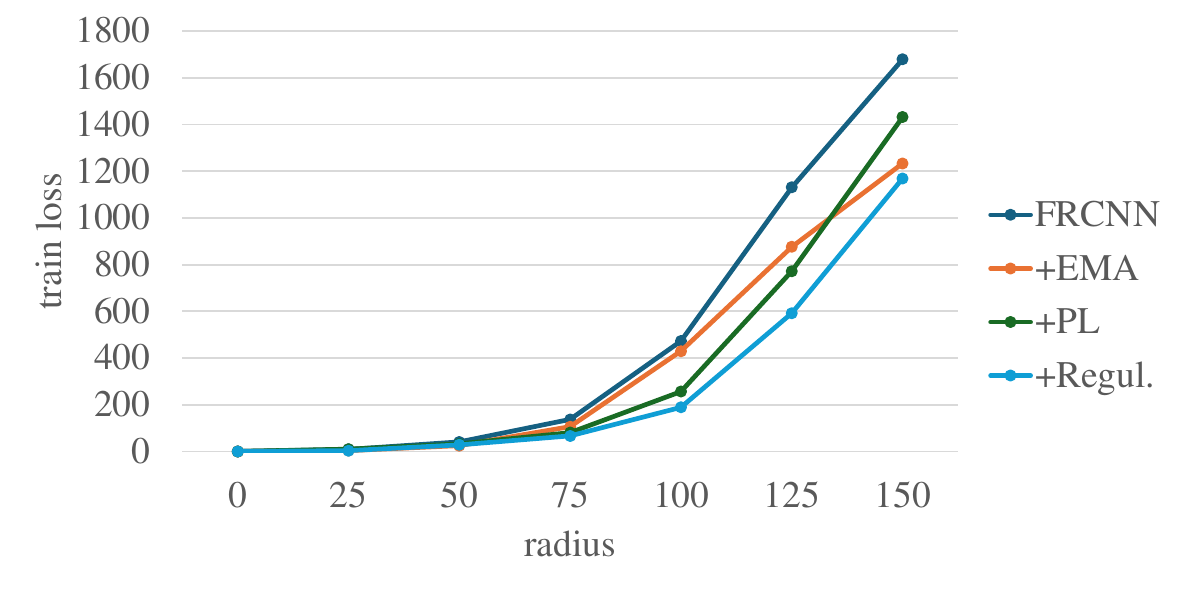}
    \subcaption{Daytime-foggy}\label{fig:flatness_train_loss_daytime_foggy}
  \end{minipage}
  \begin{minipage}[b]{0.32\hsize}
    \centering
    \includegraphics[width=\columnwidth]{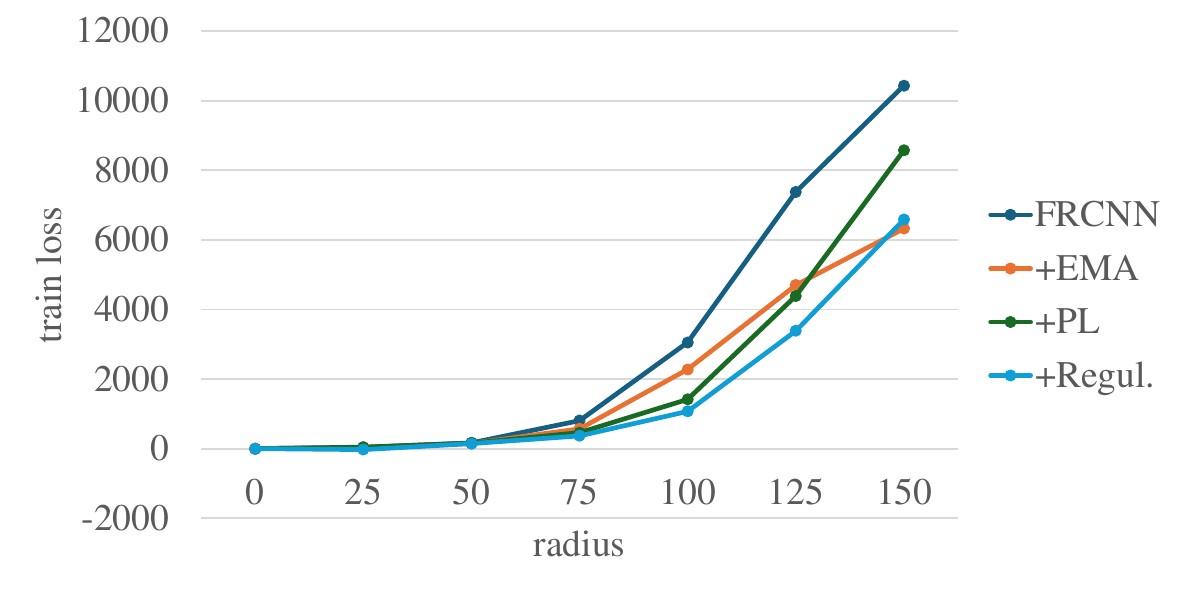}
    \subcaption{Night-sunny}\label{fig:flatness_train_loss_night_sunny}
  \end{minipage}
  \caption{Average training flatness at each training domain.}\label{fig:comp_flatness_train}
\end{figure}

\begin{figure}[h]
\centering
  \begin{minipage}[b]{0.32\hsize}
    \centering
    \includegraphics[width=\columnwidth]{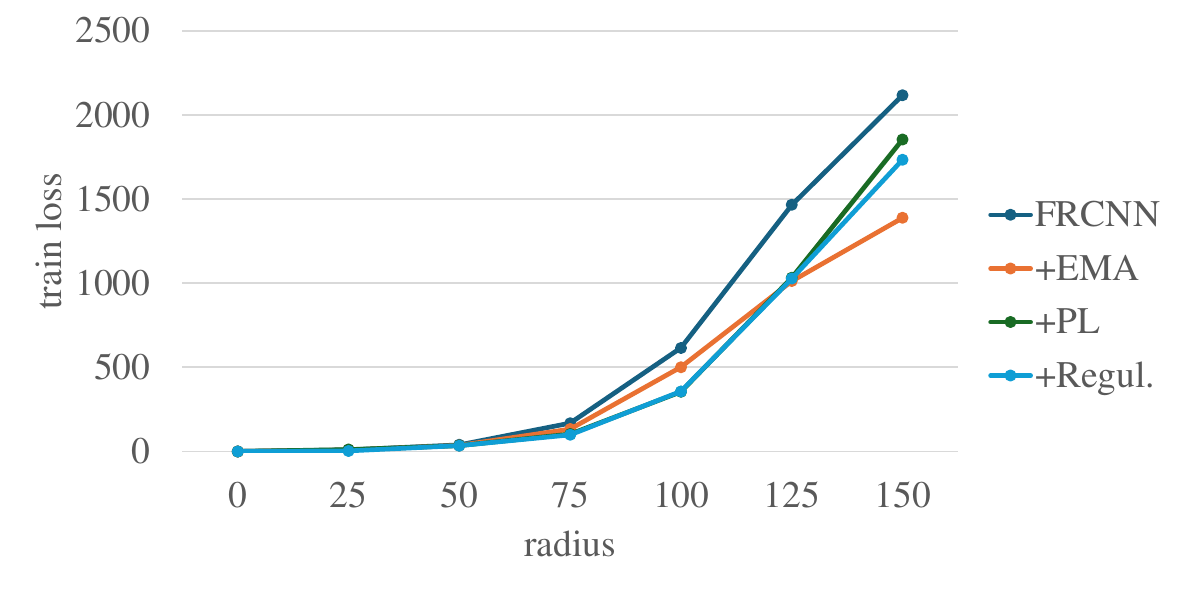}
    \subcaption{Dusk-rainy}\label{fig:flatness_test_loss_dusk_rainy}
  \end{minipage}
  \begin{minipage}[b]{0.32\hsize}
    \centering
    \includegraphics[width=\columnwidth]{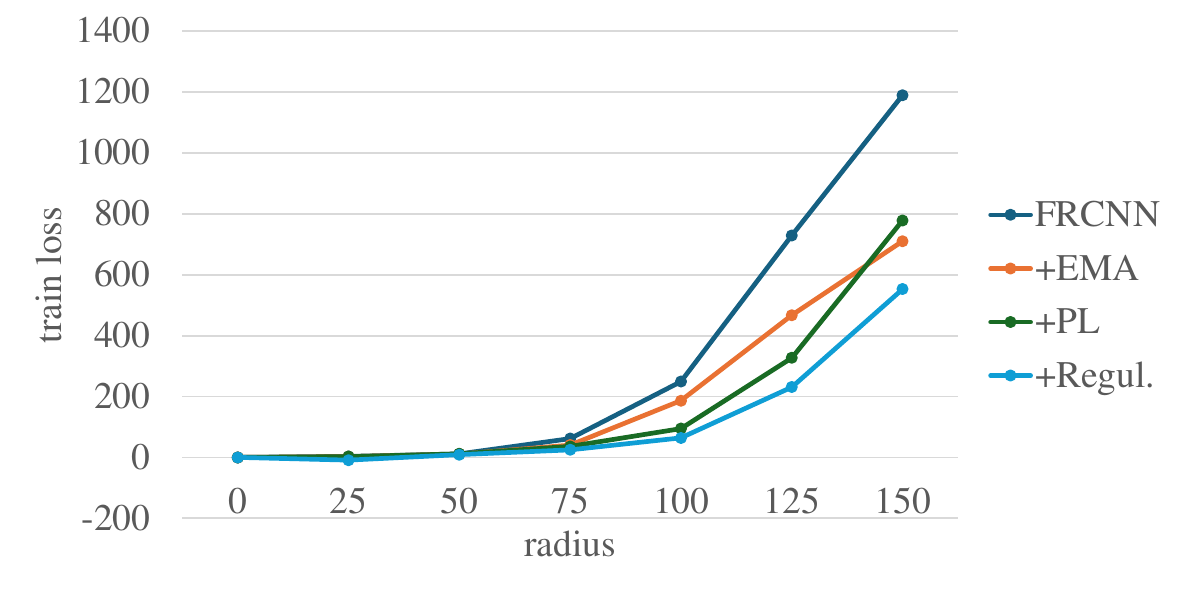}
    \subcaption{Night-rainy}\label{fig:flatness_test_loss_night_rainy}
  \end{minipage}
  \caption{Average test flatness at each target domain.}\label{fig:comp_flatness_test}
\end{figure}

Fig.~\ref{fig:comp_flatness_train} shows the average change of the training loss at each domain when perturbing the parameters ($\mathcal{F}^{\gamma}(\theta)=\mathbb{E}_{\theta'}|\mathcal{E}(\theta')-\mathcal{E}(\theta)|$ described in Sec.~\ref{sec:analysis_flatnes}), and Fig.~\ref{fig:comp_flatness_test} shows those of the test loss.
Each of EMA, PL, and the regularization lowered the changes in the losses at every domain when the radius is 125 or smaller although EMA lowered the changes the most when the radius is extremely large ($>125$).
In other words, each contributed to falling into flatter minima with a sufficiently large radius.

\subsection{Class-wise Average Precision}
Tables~\ref{tbl:comparison_dusk-rainy_perclass} and~\ref{tbl:comparison_night-rainy_perclass} show class-wise average precision on dusk-rainy and night-rainy domains, respectively.
We can see that each of EMA, PL, and regularization contributes to improving the performance on many classes except the performance drop by PL on dusk-rainy.

\begin{table*}[h]
\caption{Comparisons of AP50 at each class on dusk-rainy of the car-mounted camera dataset~\citep{wu2022single}. The values of * and ** are from~\citep{wu2022single} and~\citep{vidit2023clip}, respectively.}
\centering
{\tiny
  \begin{tabular}{llwc{6mm}wc{6mm}wc{6mm}wc{6mm}wc{6mm}wc{6mm}wc{6mm}wc{6mm}} \toprule
     setting & method & bus & bike & car & motor & person & rider & truck & mAP \\ \toprule
      Single-DGOD & FasterRCNN* & 36.8 & 15.8 & 50.1 & 12.8 & 18.9 & 12.4 & 39.5 & 26.6 \\
      Single-DGOD & CDSD*~\citep{wu2022single} & 37.1 & 19.6 & 50.9 & 13.4 & 19.7 & 16.3 & 40.7 & 28.2 \\
      Single-DGOD & CLIP-based augmentation**~\citep{vidit2023clip} & 37.8 & 22.8 & 60.7 & {\bf 16.8} & 26.8 & 18.7 & 42.4 & 32.3 \\
      Single-DGOD & Gaussian FasterRCNN & 33.9 & 14.9 & 53.6 & 4.2 & 17.4 & 13.6 & 39.2 & 25.3 \\
      Single-DGOD & Gaussian FasterRCNN + EMA & {\bf 46.3} & {\bf 24.9} & {\bf 65.9} & 11.9 & {\bf 29.1} & {\bf 23.7} & {\bf 50.0} & {\bf 36.0} \\ \bottomrule
      SS-DGOD & Gaussian FasterRCNN + EMA + PL & 40.0 & 17.3 & 61.0 & {\bf 8.0} & {\bf 23.6} & 17.1 & 45.1 & 30.3 \\
      SS-DGOD & Gaussian FasterRCNN + EMA + PL + Regul. & {\bf 40.8} & {\bf 20.1} & {\bf 61.8} & 7.8 & {\bf 23.6} & {\bf 18.3} & {\bf 46.2} & {\bf 31.2} \\ \bottomrule
      WS-DGOD & Gaussian FasterRCNN + EMA + PL & 39.0 & 19.4 & 60.4 & 9.4 & 23.8 & 17.3 & 44.0 & 30.5 \\
      WS-DGOD & Gaussian FasterRCNN + EMA + PL + Regul. & {\bf 41.7} & {\bf 22.3} & {\bf 62.1} & {\bf 11.2} & {\bf 25.3} & {\bf 18.9} & {\bf 45.9} & {\bf 32.5} \\ \bottomrule
      DGOD & Gaussian FasterRCNN & 36.2 & 18.2 & 61.3 & 7.3 & 18.4 & 15.9 & 41.9 & 28.4 \\
 \bottomrule
  \end{tabular} 
}
\label{tbl:comparison_dusk-rainy_perclass}
\end{table*}

\begin{table*}[h]
\caption{Comparisons of AP50 at each class on night-rainy of the car-mounted camera dataset~\citep{wu2022single}. The values of * and ** are from~\citep{wu2022single} and~\citep{vidit2023clip}, respectively.}
\centering
{\tiny
  \begin{tabular}{llwc{6mm}wc{6mm}wc{6mm}wc{6mm}wc{6mm}wc{6mm}wc{6mm}wc{6mm}} \toprule
     setting & method & bus & bike & car & motor & person & rider & truck & mAP \\ \toprule
      Single-DGOD & FasterRCNN* & 22.6 & 11.5 & 27.7 & 0.4 & 10.0 & 10.5 & 19.0 & 14.5 \\
      Single-DGOD & CDSD*~\citep{wu2022single} & 24.4 & 11.6 & 29.5 & {\bf 9.8} & 10.5 & {\bf 11.4} & 19.2 & 16.6 \\
      Single-DGOD & CLIP-based augmentation**~\citep{vidit2023clip} & 28.6 & {\bf 12.1} & 36.1 & 9.2 & {\bf 12.3} & 9.6 & 22.9 & 18.7 \\
      Single-DGOD & Gaussian FasterRCNN & 20.4 & 7.7 & 31.0 & 0.5 & 6.8 & 5.6 & 21.3 & 13.3 \\
      Single-DGOD & Gaussian FasterRCNN + EMA & {\bf 33.9} & 11.1 & {\bf 38.5} & 0.8 & 10.5 & 8.8 & {\bf 29.2} & {\bf 19.0} \\ \bottomrule
      SS-DGOD & Gaussian FasterRCNN + EMA + PL & 35.7 & 9.8 & {\bf 46.7} & 1.4 & 12.6 & 10.8 & {\bf 32.0} & 21.3 \\
      SS-DGOD & Gaussian FasterRCNN + EMA + PL + Regul. & {\bf 37.0} & {\bf 10.3} & 46.3 & {\bf 2.8} & {\bf 12.9} & {\bf 12.0} & 31.8 & {\bf 21.9} \\ \bottomrule
      WS-DGOD & Gaussian FasterRCNN + EMA + PL & {\bf 38.6} & 11.3 & {\bf 47.9} & {\bf 2.9} & 13.4 & 11.2 & {\bf 32.1} & 22.5 \\
      WS-DGOD & Gaussian FasterRCNN + EMA + PL + Regul. & 38.3 & {\bf 13.4} & 46.2 & 2.7 & {\bf 15.1} & {\bf 14.0} & 32.0 & {\bf 23.1} \\ \bottomrule
      DGOD & Gaussian FasterRCNN & 38.9 & 7.6 & 46.7 & 1.8 & 9.8 & 11.3 & 32.1 & 21.2 \\
 \bottomrule
  \end{tabular} 
}
\label{tbl:comparison_night-rainy_perclass}
\end{table*}

\subsection{Qualitative Results}
\begin{figure*}[h]
    \centering
    \begin{minipage}{\linewidth}
    \centering
    \includegraphics[width=\linewidth]{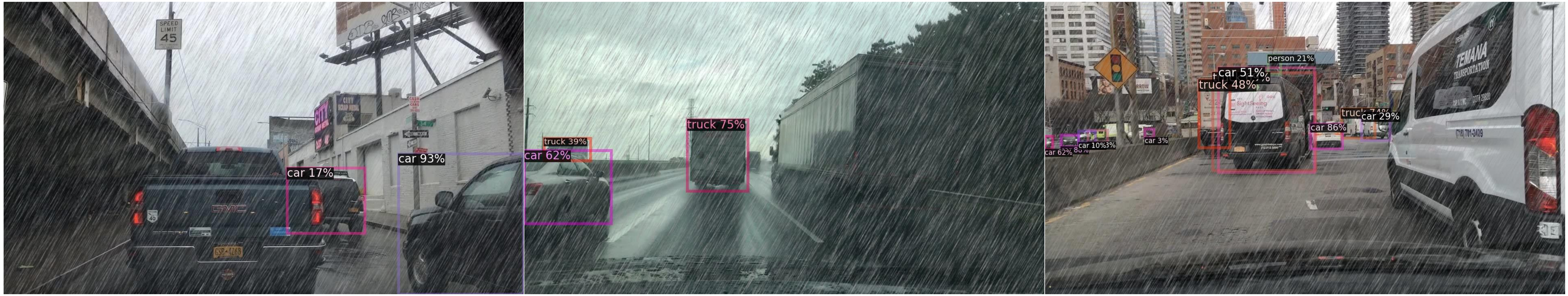}
    \subcaption{Gaussian FasterRCNN trained on Single-DGOD setting (i.e., labeled data on daytime-sunny).}
    \end{minipage}
    \begin{minipage}{\linewidth}
    \centering
    \includegraphics[width=\linewidth]{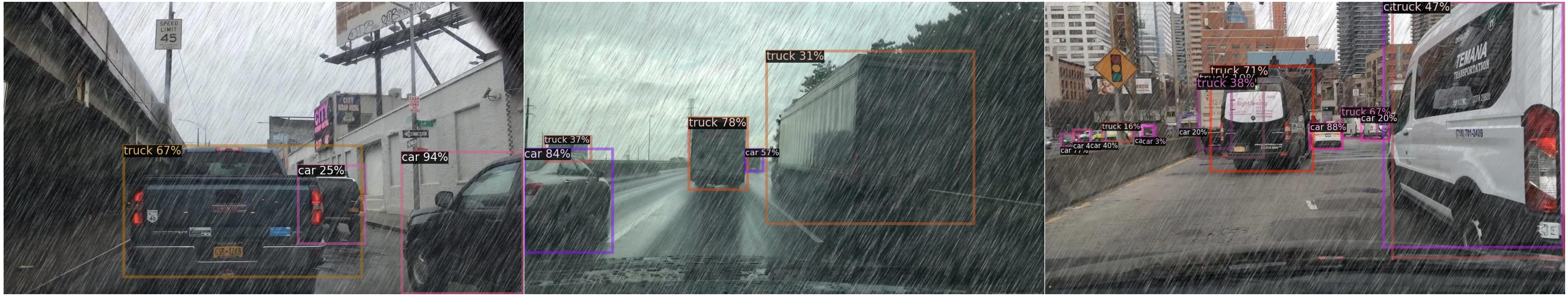}
    \subcaption{Gaussian FasterRCNN + EMA + PL + Regul. trained on SS-DGOD setting (i.e., labeled data on daytime-sunny and unlabeled data on night-sunny and daytime-foggy).}
    \end{minipage}
    \caption{Qualitative comparisons on dusk-rainy.}
    \label{fig:qualitative_comp_dusk_rainy}
\end{figure*}

\begin{figure*}[h]
    \centering
    \begin{minipage}{\linewidth}
    \centering
    \includegraphics[width=\linewidth]{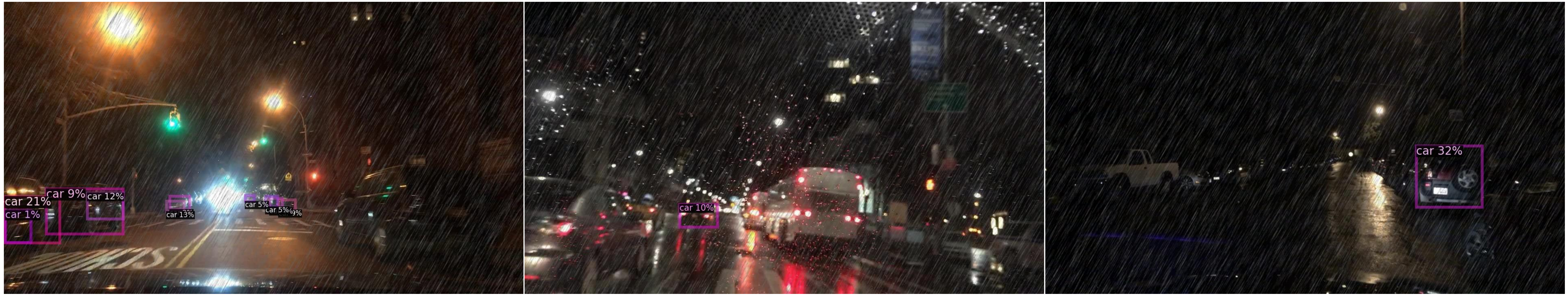}
    \subcaption{Gaussian FasterRCNN trained on Single-DGOD setting (i.e., labeled data on daytime-sunny).}
    \end{minipage}
    \begin{minipage}{\linewidth}
    \centering
    \includegraphics[width=\linewidth]{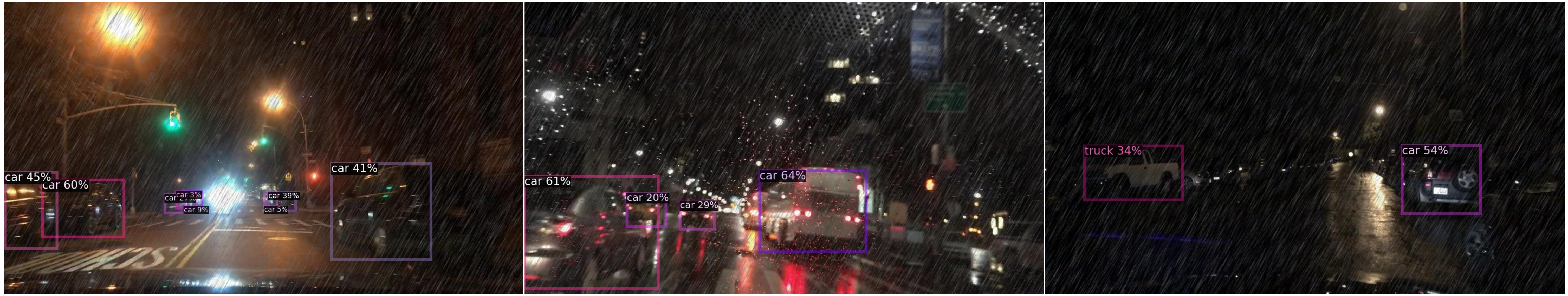}
    \subcaption{Gaussian FasterRCNN + EMA + PL + Regul. trained on SS-DGOD setting (i.e., labeled data on daytime-sunny and unlabeled data on night-sunny and daytime-foggy).}
    \end{minipage}
    \caption{Qualitative comparisons on night-rainy.}
    \label{fig:qualitative_comp_night_rainy}
\end{figure*}

Figs.~\ref{fig:qualitative_comp_dusk_rainy} and~\ref{fig:qualitative_comp_night_rainy} show the qualitative comparison on dusk-rainy and night-rainy, respectively. 
Similar to the artistic image dataset, the baseline model had false negative detections, which were improved by EMA, PL, and regularization.

\section{Training Details}\label{sec:training_details}
On the artistic style image dataset, the detectors were trained with 10,000 and 20,000 iterations for the pretraining and the student-teacher learning of SS-DGOD (or WS-DGOD), respectively.
During the training, we saved the models and evaluated the performance on the validation at every 2,000 iterations, and the best model was used for the evaluation.
The whole training took about one day.
For fair comparisons, the compared models on Single-DGOD and DGOD were trained with 30,000 iterations, and the best models at the validation of every 2,000 iterations were used for evaluation.

On the car-mounted camera dataset, we performed the same procedure for training, validation, and evaluation, but the numbers of iterations for the pretraining and the student-teacher learning were set to 20,000 and 40,000 respectively, and the validation was conducted at every 4,000 iterations.
The whole training took about two days.
For fair comparisons, the compared models on Single-DGOD and DGOD were trained with 60,000 iterations, and the best models at the validation of every 4,000 iterations were used for evaluation.

\section{More discussions}

\subsection{The Other Semi-supervised Domain Generalization Setting}
There are two types of settings on semi-supervised domain generalization. The first setting assumes that only a part of the samples in each domain are labeled~\citep{zhou2023semi}.
The other one assumes that only a part of the source domains are labeled~\citep{LIN2024110280}.
In this paper, we tackled the second setting for simplicity.
To confirm the effectiveness of the Mean Teacher framework and the regularization on the other setting is one of our future works.

\subsection{Broader Impacts}\label{sec:impacts}
In this work, we tackled the task of semi-supervised and weakly-supervised domain generalization for object detection (SS-DGOD and WS-DGOD), which are more practical settings than previous works.
Also, we showed the good performance of the Mean Teacher learning framework, its interpretations, and a simple regularization method to boost the performance. Therefore, we believe that this paper has a potential positive social impact to enable practitioners or researchers to train robust object detectors to unseen domains in a simpler way than previous approaches.
In addition, because Mean Teacher has been used across various tasks, our novel interpretation of why Mean Teacher becomes robust to unknown domains is likely to have a broad impact across a wide range of tasks.
We are unable to identify any pertinent information concerning potential negative impacts.

\end{document}